\documentclass[sigconf]{acmart}

\usepackage{booktabs} 

\usepackage{array}
\usepackage[linesnumbered,ruled]{algorithm2e}
\usepackage{setspace}

\usepackage{exercise}
\usepackage{lipsum}

\usepackage{amsmath,amsfonts,bm}

\def\eqref#1{equation~\ref{#1}}

\def\1{\bm{1}}

\DeclareMathAlphabet{\mathsfit}{\encodingdefault}{\sfdefault}{m}{sl}
\SetMathAlphabet{\mathsfit}{bold}{\encodingdefault}{\sfdefault}{bx}{n}

\newcommand{\R}{\mathbb{R}}

\DeclareMathOperator*{\argmin}{arg\,min}

\setcopyright{rightsretained}

\acmDOI{10.475/123_4}

\acmISBN{123-4567-24-567/08/06}

\acmConference[SIGKDD'19]{ACM SIGKDD conference}{August 2019}{Anchorage, Alaska, USA}
\acmYear{2019}
\copyrightyear{2019}

\acmArticle{4}
\acmPrice{15.00}

\begin{document}
\title{NeuralWarp: Time-Series Similarity with Warping Networks}

\author{Josif Grabocka}
\affiliation{%
  \institution{Information Systems and Machine Learning Lab}
  \streetaddress{Samelsonplatz 22}
  \city{31141 Hildesheim}
  \state{Germany}
  \postcode{31135}
}
\email{josif@ismll.uni-hildesheim.de}

\author{Lars Schmidt-Thieme}
\affiliation{%
  \institution{Information Systems and Machine Learning Lab}
\streetaddress{Samelsonplatz 22}
\city{31141 Hildesheim}
\state{Germany}
\postcode{31135}
}
\email{schmidt-thieme@ismll.uni-hildesheim.de}

\renewcommand{\shortauthors}{Grabocka et al.}

\begin{abstract}
	
Research on time-series similarity measures has emphasized the need for elastic methods which align the indices of pairs of time series and a plethora of non-parametric measures have been proposed for the task. On the other hand, deep learning approaches are dominant in closely related domains, such as learning image and text sentence similarity. In this paper, we propose \textit{NeuralWarp}, a novel measure that models the alignment of time-series indices in a deep representation space, by modeling a warping function as an upper level neural network between deeply-encoded time series values. Experimental results demonstrate that \textit{NeuralWarp} outperforms both non-parametric and un-warped deep models on a range of diverse real-world datasets.
	
\end{abstract}

\keywords{Deep Learning, Time series, Similarity Learning}

\maketitle
	
\section{Introduction}
\label{sec:introduction}

Time-series similarity is one of the crucial problems for the machine learning community and has been extensively researched in the recent decades. The predominant approaches in learning (diss)similarity measures have stressed the need for elastic functions in aligning/warping the indices of time series. Such a warping attribute of a similarity function is essential given the high degree of intra-class variations in time-series instances, exhibited in the form of shifts, distortions, noise and diverse scalings of patterns. For this reason, Dynamic Time Warping (DTW) - a method that aligns the indices of two series - has been shown to perform accurately in tackling intra-class series variations~\cite{sakoe_dynamic_1978,Ding:2008:QMT:1454159.1454226}. DTW aligns series in a way that the total sum of distances between the corresponding values of the aligned indices is the smallest possible~\cite{SakoeChiba71}. The community on time-series similarity research have proposed a plethora of elastic measures for rivaling DTW, such as the Time Warp Edit Distance, which captures the elasticity aspect by borrowing the concept of edit distances from string similarity~\cite{marteau_time_2009}. The unifying characteristic of the prior work is the fact that those methods are static (a.k.a. non-parametric), meaning they can operate without having to train the respective similarity function through supervised learning. On the contrary, similarity measures in closely-related application domains, such as image or text mining, rely heavily on exploiting parametric models in the form of deep neural networks. The mechanism of learning the similarity between series via neural networks is centered on the Siamese architectures, which is a metaphor for using the same Deep Learning models to produce the latent/deep embedding of a pair of time series~\cite{Bromley:1993:SVU:2987189.2987282}. Typically, such latent embeddings are computed through Convolutional Neural Networks (CNN) in the realm of images~\cite{hu_discriminative_2014,yi_deep_2014}, or Recurrent Neural Networks (RNN) in the case of text~\cite{mueller_siamese_2016}. The distance between time series is therefore usually formalized as the $L_p$ norm distance between the latent embedding tensors in the deep representation space~\cite{mueller_siamese_2016, pei_modeling_2016}. 

We emphasize that deep similarity measures are not thoroughly explored for general time-series data. In particular, Siamese architectures do not directly model the elastic alignment of time-series instances in the deep representation space. In this paper, we close the gap between the research streams in elastic time-series measures and the deep Siamese architectures by proposing a novel technique that directly models the alignment of time-series representations in a deep neural space. Our motivation is based on the fact that optimal alignment paths can be re-formulated as an all-pairs distance with a warping indicator function. Furthermore, we argue that such a function cannot be numerically trained directly on raw time series measurements, because raw values provide an insufficient level of dis-ambiguity contexts (Section~\ref{sec:paramwarpingsim}).

As a result, we propose to project raw time series into a latent representation by means of encoder neural networks, such as CNN or RNN, in a way that indices represent the context of the pattern at a particular time index, instead of the raw value. Afterwards, we introduce a parametric alignment function for warping the encoded contexts of the time-series instances. Such a warping function is an upper layer neural network that takes as input the encoded deep contexts of a pair of index values from two time series, and outputs the probability of aligning those indices. Therefore, the $L_p$ norm distance of standard Siamese architectures is transformed into an elastic measure for capturing intra-class variations of time-series patterns. The novel deep alignment measure is named NeuralWarp and is detailed in Section~\ref{sec:deepwarp}. We propose a loss function that maximizes the similarity of the warped deep measure of given pairs of similar series, while minimizing the similarity of pairs of dissimilar pairs of series. The optimization of the loss is carried out by a first-order minimization in Section~\ref{sec:optimization}. 

We conducted extensive experiments to test the accuracy of the proposed elastic deep similarity by comparing it against strong non-parametric methods, as well as un-warped deep Siamese architectures. We noticed that a warper function significantly improves the performance of existing Siamese approaches, in terms of helping a Nearest Neighbor classifier achieve higher accuracies. Moreover, the experimental results of Section~\ref{sec:results} demonstrate that NeuralWarp is superior in accuracy compared to strong non-parametric baselines. Our results show that encoders based on Recurrent Neural Networks (RNN) achieve the best performance in providing contextual features to the proposed warping function.  In addition, we compared our similarity measure against same-capacity deep architectures that are trained directly for classification and found out that the warped RNN models are more accurate than classification RNN.

\section{Related Work}
\label{sec:relatedwork}

Since the genesis of Recurrent Neural Networks, architectures that handle the time shifting aspects in signal data have been proposed~\cite{sun_time_1992}. In strong contrast to other domains, such as speech and image mining, the similarity measures within the time series domain have been mainly focused on non-parametric functions, such as DTW, TWED, etc. In fact, deep learning became out of fashion in the time-series research community, primarily in favor of elastic similarity measures~\cite{Bagnall2017}. Yet, a recent paper showed that relatively off-the-shelve, yet properly tuned, deep networks with batch normalization layers and global average pooling can achieve state-of-the-art classification accuracy~\cite{wang_time_2017}. Throughout this section we are briefly covering the prior published work on similarity learning for time series data.

\subsection{Similarity Learning}

Similarity learning is an established task for the machine learning community and has been extensively covered in the recent decade. For instance, in the context of image retrieval, similarity learning is used to 'search by image' in a search engine application~\cite{wang_learning_2014}. In the music information retrieval community, the similarity between recorded audio data is crucial for identifying genres and/or composers~\cite{li_content_2004, mcfee_learning_2012}. Moreover, for the bio-metric identification realm it is important to learn the similarity between individuals based on recorded video signals~\cite{zhu_deep_2018}. Last, but not least, the semantic similarity between text contents is essential for diverse tasks in computational linguistics~\cite{mueller_siamese_2016}. 

Recently, deep learning approaches have become the de-facto standard in image retrieval systems~\cite{wan_deep_2014}. Deep similarity measures are learned by \textbf{Siamese} networks, a metaphor that depicts using the same network for deriving the latent representation of a pair of instances~\cite{Bromley:1993:SVU:2987189.2987282}. The Siamese mechanism computes either the L1/L2 distance \cite{mueller_siamese_2016,DBLP:conf/cvpr/ZagoruykoK15, hu_discriminative_2014}, or the cosine similarity \cite{yi_deep_2014} between the pairs of instances. Not every dissimilarity measure is strictly speaking a distance metric, since metrics have to obey three conditions, such as non-negativity, symmetry, triangle inequality and the identity of indiscernibles. Learning a distance metric in a latent representation was initially optimized for clustering~\cite{xing2003distance}. In the context of sequences, a metric learning approach for time sequence optimizes the temporal sequence alignment in form of a Mahalanobis distance~\cite{garreau_metric_2014}.

\subsection{Time-series Similarity}

Learning the similarity of time series has been long conducted using static (i.e. non-parametric) measures such as Euclidean distance and Dynamic Time Warping. In strong contrast to other related domains, such as speech and image mining, deep learning similarity measures have so far not been thoroughly explored for generic multivariate time series data. Among static measures, a series of those approaches have been inspired by the edit distance concept of strings, i.e. the number of operations needed to convert one string to another. In that regards, the Edit Distance with Real Penalty has first adapted this concept for time-series similarity~\cite{Chen:2004:MLE:1316689.1316758}. Recently, another method combines the aspects of edit distance with time warping. \cite{marteau_time_2009}. Another technique learns the longest common sub-sequences between time series as a notion of similarity~\cite{Hunt:1977:FAC:359581.359603}. 

The most notorious (dis)similarity measure is Dynamic Time Warping (DTW)~\cite{SakoeChiba71, sakoe_dynamic_1978}, which computes the alignment of two series' indices. An initial empirical survey concluded that DTW is a very accurate measure in terms of classification accuracy~\cite{Ding:2008:QMT:1454159.1454226}. DTW computes the best alignment between indices of two time series, in a way that the sum of aligned series values is the smallest possible. Numerous papers have elaborated lateral aspects of DTW, while its $\mathcal{O}(n^2)$ computational run-time has been ameliorated by various schemes, such early abandoning and lower bounds~\cite{Rakthanmanon:2012:SMT:2339530.2339576}. A multi-level decomposition of the series regions can reduce the computational time and space that DTW requires to a linear complexity~\cite{salvador_toward_2007}. Furthermore, another paper aims at deriving a differentiable variant of DTW that can be used for numerical optimizations of loss functions~\cite{cuturi_soft-dtw:_2017}. In comparison to the best warping alignment strategy of DTW, a prior work proposes to use the soft-minimum of all alignment paths, yielding a mathematically-sound distance metric~\cite{cuturi_fast_2011}. Recently, the idea of all-pairs similarity join has attracted attention in the time series research community~\cite{7837992}. Moreover, this representation has been applied to measure time series similarity as the degree of common sub-sequences~\cite{mpdist}. It is worth pointing out that the most recent and empirically exhausting survey of similarity measures for time series concluded that Dynamic Time Warping (DTW) and Time Warped Edit Distance (TWED) are the most competitive options~\cite{serra_empirical_2014}. In this paper we empirically compare the proposed elastic similarity measure against both DTW and TWED. 

\subsection{Deep Learning Similarity for Time Series} 

Time series are mainly modeled through Recurrent Neural Networks (RNN), or Convolutional Neural Networks (CNN). With regards to utilizing RNN for sequences, it has been indicated that Long Short Term Memory (LSTM) has the property of being invariant to pattern shifts in the time domain~\cite{tallec_can_2018}, and are considered a natural model for time series. Furthermore, CNN have the ability to capture discriminative series patterns and can be considered a generalization of shapelet-based classifiers~\cite{wang_time_2017,Grabocka:2014:LTS:2623330.2623613}. Combinations of both models have also been proposed, in the form of LSTM fully convolutional networks~\cite{fazle_lstm_2017}. In the context of capturing similarity, the Siamese architecture was initially proposed for detecting similar handwritten signatures~\cite{bromley1994signature}. In that perspective, a recent work aims at building a Siamese network for extracting latent representations of time series, in a way that the similarity metric considers a global alignment function in the deep space~\cite{Che2017DECADEA}. During the same year, two separate papers have further utilized Siamese RNN networks for capturing the similarities of sequences in the domains of text and time series~\cite{mueller_siamese_2016, pei_modeling_2016}. Another recent work, has explored a similar architecture in the realm of action recognition~\cite{roy_action_2018}. 
In this paper we further extend Siamese deep networks to warp time series in the latent/deep representation, by introducing a warping network that contextualizes the alignment of indices in the latent space.

\subsection{Novelty and Research Hypothesis}
\label{sec:novelty}

The motivation of this paper is based on two key observations:

\begin{enumerate}
	\item Deep learning is not \textit{thoroughly} explored for optimizing time-series similarity measures;
	\item The warping aspect of time-series similarity is not \textit{directly} modeled in deep networks.
\end{enumerate}

Whilst deep learning offers the best family of models for harvesting the power of machine learning, there is still no direct and efficient way to align time-series in a deep representation. The optimal warping alignment utilized by the Dynamic Time Warping measure is not directly usable in a latent representation, because, despite attempts to provide mathematically smooth relaxations~\cite{cuturi_soft-dtw:_2017}, it is still not a differentiable function. On the other hand, Recurrent Neural Networks are not an optimal mechanism to model time series, too. RNN can produce a cumulative representation for each time index, however they do not define how the indices of two time series should be aligned.

In this paper, we propose \textbf{NeuralWarp:} the first elastic deep learning measure for time-series similarity. Our suggestion is to model an alignment/warping path between two sequences' values as a dedicated neural network function in the deep representation of the sequence values. In that way, we jointly learn a deep embedding of the time series (with RNN, or CNN models) together with a warping neural networks that learns to align values in the latent space. Following this line of thought, the primary objective of this paper is to address the following research hypothesis:

\begin{itemize}
	\item {Can parametric warping functions trained in a deep representation of time series yield \textit{better}\footnote{Compared through the classification accuracy of a Nearest Neighbor classifier using rivaling similarity measures.} similarity measures compared to:}
	\begin{itemize}
		\item[a)] \textit{Non-parametric warping measures (e.g. DTW, TWED)},  \underline{and}
		\item[b)] \textit{Un-warped deep variants (e.g. Siamese CNN or RNN)} ?
	\end{itemize}
\end{itemize}

\section{Alignment Paths}

Even though DTW is a well-known algorithm, we would still invest some lines to explain it, because the idea of optimal alignment path is crucial for further understanding the remaining sections of the paper. DTW learns the best alignment path between two time series $A \in \R^{T \times D}$ and $B \in \R^{T \times D}$ with $T$ measurements of $D$ channels each, as the list of index pairs $\left(\phi_{k}^{(A)}, \phi_{k}^{(B)}\right), \phi \in \left\{1,\dots,T\right\}$. Equation~\ref{eq:warppathdistance} formalizes the concept behind an optimal warping path. 

\begin{eqnarray}
\label{eq:warppathdistance}
\mathcal{D}{(A,B)} &:=& \min\limits_{\phi} \sum_{k=1}^{|\phi|} {||A_{\phi_{k}^{(A)}} - B_{\phi_{k}^{(B)}}||}_2^2 \\
\label{eq:monotonicallyconstraint}
\begin{pmatrix}
\phi_{k}^{(A)} - \phi_{k-1}^{(A)} \\
\phi_{k}^{(B)} - \phi_{k-1}^{(B)}
\end{pmatrix} &:=& \left\{ \begin{pmatrix} 0 \\ 1 \end{pmatrix}, \begin{pmatrix} 1 \\ 0 \end{pmatrix}, \begin{pmatrix} 1 \\ 1 \end{pmatrix}  \right\}
\end{eqnarray}

\begin{figure}[h]
	\centering
	\includegraphics[width=0.9\linewidth,trim=1.5cm 3.3cm 0cm 1.8cm, clip=true]{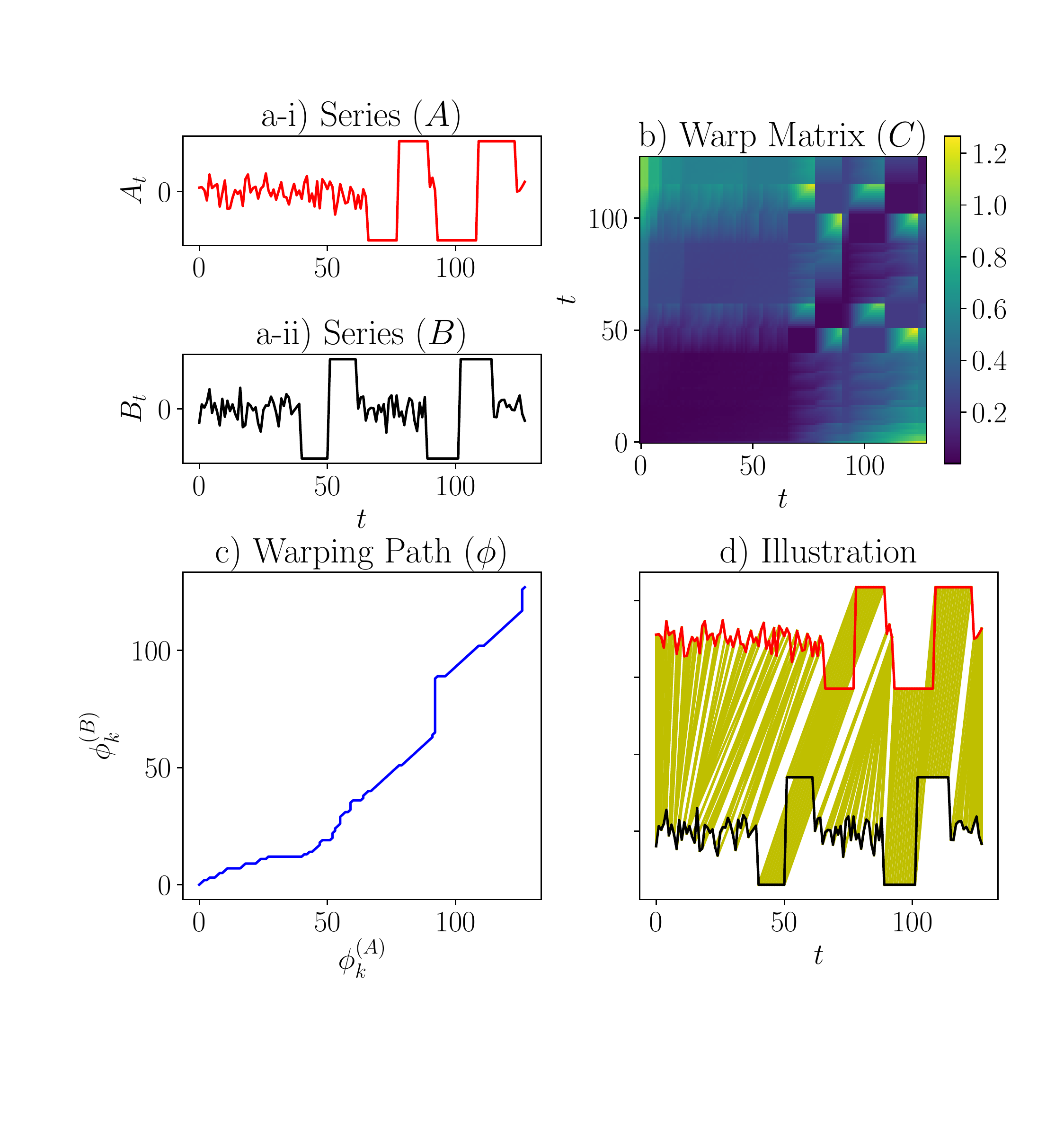}
	\caption{A DTW illustration with series from the "Two Patterns" dataset~\cite{UCRArchive2018}, where the warping path $\phi$ is derived from the warping matrix $C$.}
	\label{fig:warping}
\end{figure}

In the case of DTW the index pairs $\phi$ should be monotonically non-decreasing. Figure~\ref{fig:warping} provides an illustration of how DTW operates on two illustrated series. The warping path $\phi$ is shown on subplot c) and the index pairs are depicted on d).

The Dynamic Time Warping distance is solved by a recursive dynamic programming approach, which is formalized by Equation~\ref{eq:warpingmatrix}. The total distance of the best warping path between the first $i$ indices of the first series $A_{1:i,:}$ and the first $j$ indices of the second series $B_{1:j,:}$ is defined as $C_{i,j}$, where $C$ often called the warping matrix in the time-series literature.

\begin{eqnarray}
\label{eq:warpingmatrix}
D(A,B) &=& C_{T,T}, \;\; \text{ where: } \\
C_{i,j} &=& ||A_i-B_j||_2^2 + \min\left\{C_{i-1,j}, C_{i,j-1}, C_{i-1,j-1}\right\}
\end{eqnarray}

\section{Parametric Warping Similarity}
\label{sec:paramwarpingsim}

It is possible to rewrite the Dynamic Time Warping distance of Equation~\ref{eq:warppathdistance} as the Equations~\ref{eq:unwarpeddistance}-\ref{eq:warpingfunction}. In other words, we can represent the optimal alignment of Equation~\ref{eq:warppathdistance} as a measure between all the indices of both series, by multiplying it with an indicator function $\Phi$ which determines whether the value pairs $(A_i, B_j)$ should be matched, and checking whether the index $(i,j)$ is part of an optimal alignment path similar to the one computed through Equation~\ref{eq:warpingmatrix}.

\begin{eqnarray}
\label{eq:unwarpeddistance}
\mathcal{D}{(A,B)} &=& \sum_{i = 1}^{T} \; \sum_{j = 1}^{T}  {|| A_i - B_j ||}_{2}^{2} \;\; \Phi{\left( A_i, B_j; \; \phi \right)} \\ 
\label{eq:warpingfunction}
\Phi{\left(A_i, B_j; \; \phi\right)} &=& \begin{cases}
1 & (i,j) \in \phi \\
0 & (i,j) \notin \phi \\
\end{cases}
\end{eqnarray}

Nevertheless, such a conversion can hint an important question: \textit{Can we actually define and learn such a warping function $\Phi: \R \times \R \rightarrow \left[0,1\right]$ in a principled (supervised) manner?} Before following this line of thinking, one should notice that DTW is optimal with respect to achieving the sum of distance values of the aligned index pairs. However, DTW is a static measure that ignores the peculiar characteristics of a dataset. In other words, it cannot ignore a part of series and consider only certain sub-segments that are relevant for capturing the similarity, due to the monotonically non-decreasing constraint of Equation~\ref{eq:monotonicallyconstraint}. Therefore, it is important to optimize alignments in a supervised manner that trains a warping path for each particular dataset. 

The obvious dilemma is how to parameterize and learn the warping alignment function $\Phi$? The naive approach is to provide as input the values of the time series and output a binary indicator representing whether the values should be aligned: $\Phi(A_{i}, B_{j}): \R \times \R \rightarrow [0,1]$. Obviously, one can model the function as a neural network with two inputs (series values) and an output neuron with a Sigmoid activation function that predicts the alignment probability. However, the series values corresponding to the indices alone provide an insufficient level of information for deciding whether or not to align.

\begin{figure}[h]
	\centering
	\includegraphics[width=0.9\linewidth,trim=3cm 1cm 8cm 1.3cm, clip=false]{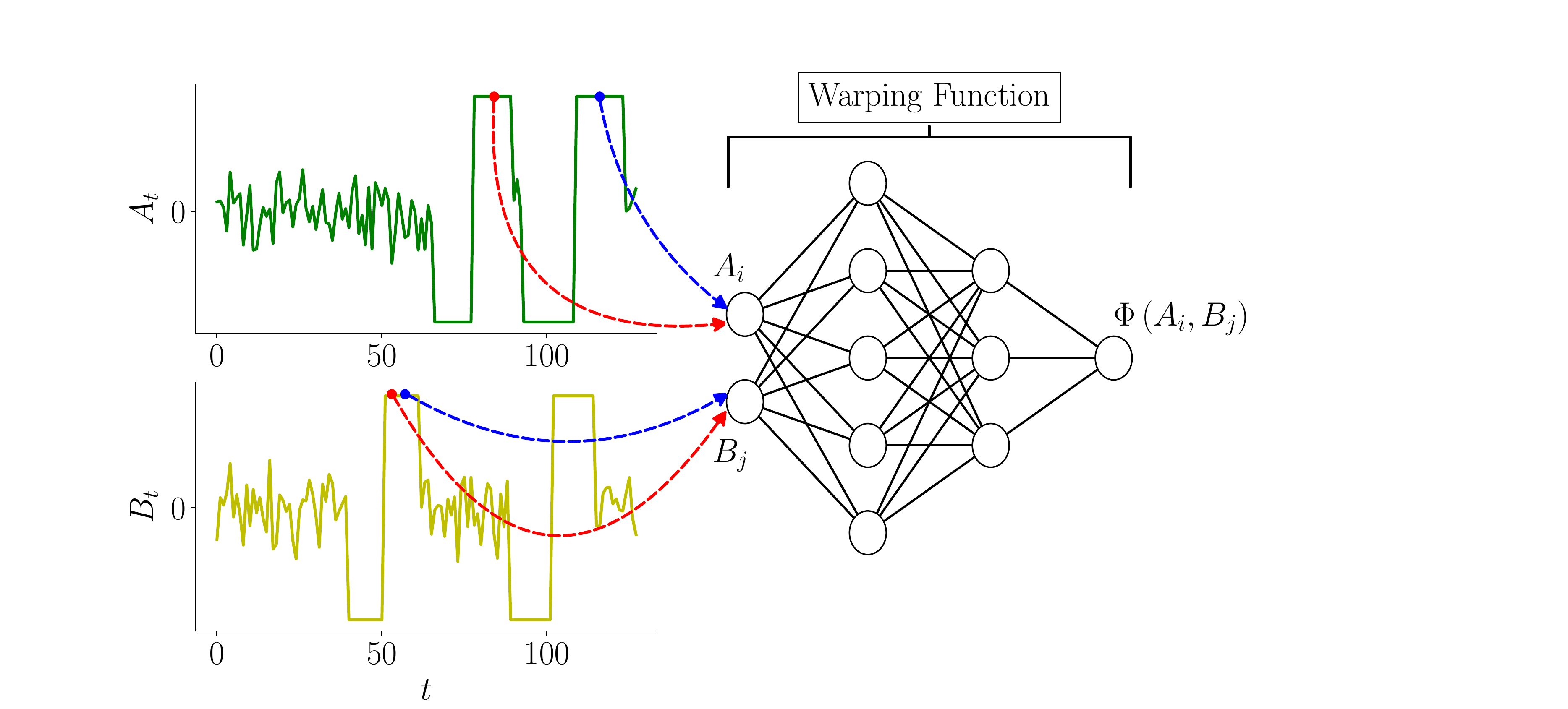}
	\caption{A warping function defined directly on time-series values fails to capture the "warping context" in cases where input pairs have the same values.}
	\label{fig:warpingcontext}
\end{figure}

Figure~\ref{fig:warpingcontext} illustrates the issue arising from training a warping function directly on time-series values. The two illustrated series are the same as those in Figure~\ref{fig:warping} where the pair of values indicated by red lines should be matched, while the pair of values indicated by blue lines should not. However, those two pairs of time indices have the same values. Therefore, a neural network, which is a deterministic function, cannot compute different warping alignment outputs when given the same inputs. The crucial problem of this approach is that an efficient warping function should be inputted the "context" of an index value within the time-series, instead of merely the value. In that aspect, Dynamic Time Warping uses the aggregated optimal distance so far (i.e. warping matrix $C$ from Equation~\ref{eq:warpingmatrix}) as the context information. 

\subsection{NeuralWarp}
\label{sec:deepwarp}

%

\begin{figure*}[ht!]
	\centering
	\includegraphics[width=0.9\linewidth,trim=0cm 1cm 4cm 2.5cm, clip=true]{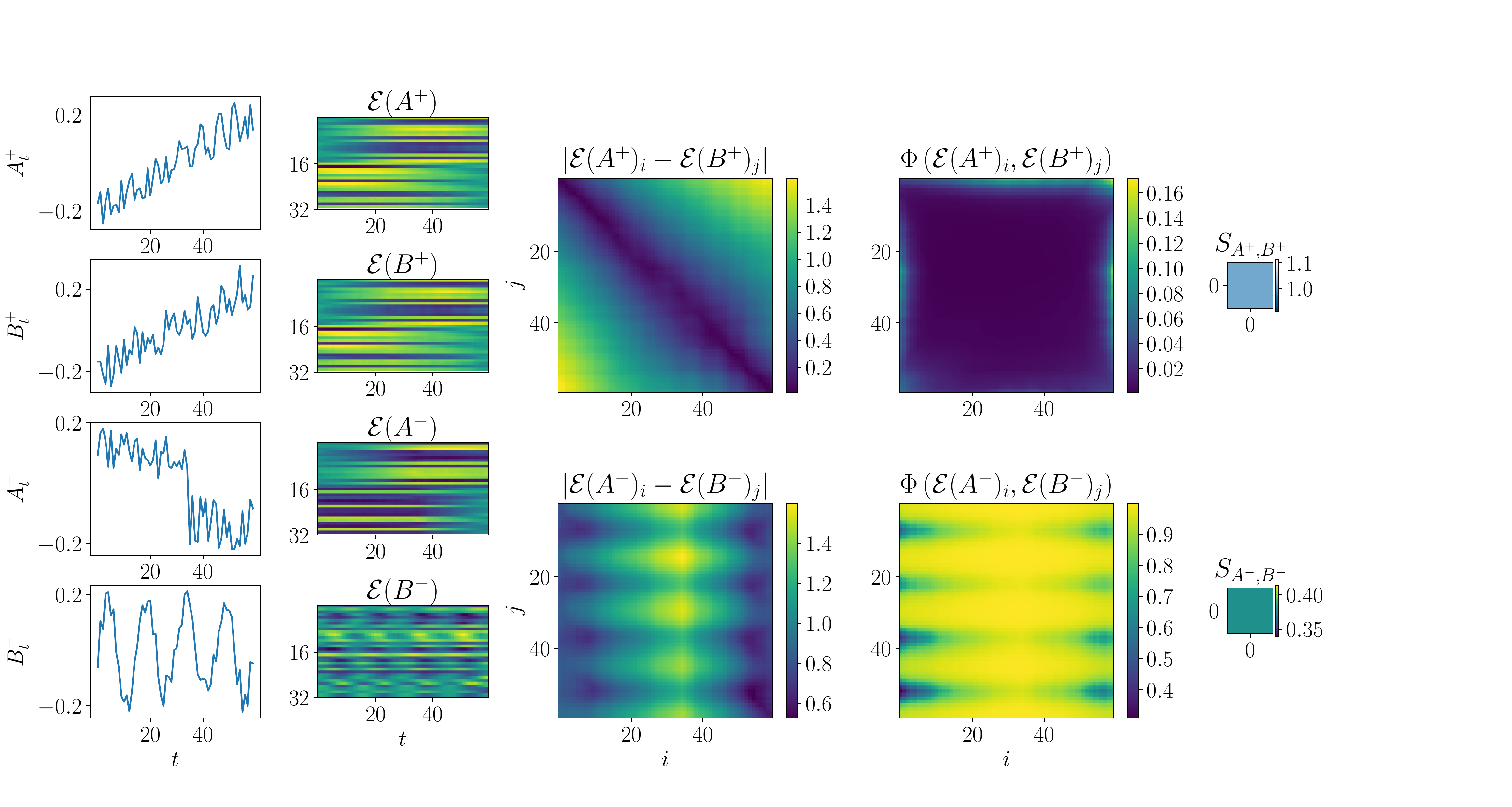}
	\caption{A bi-directional RNN encoding network $\mathcal{E}$ with one layers having $[16]$ cells and a warper $\Phi$ network of size $[16, 8, 4, 1]$ trained for 10K batches of 30 pairs on the univariate "Synthetic Control" dataset~\cite{UCRArchive2018}. The image illustrates the encoding and warping of a similar $A^{+}, B^{+}$ and a dissimilar $A^{-}, B^{-}$ pairs of univariate series.}
	\label{fig:mechanism}
\end{figure*}

In order to mine the information encapsulated by the context at a particular index of a time-series, we propose to convert the time series to a new representation by means of an encoder function. The function $\mathcal{E}: \R^{T \times D} \rightarrow \R^{T \times K}$ will convert each of the $D$-dimensional (multi-channel) measurements of a time series to a $K$-dimensional context vector. There are two standard deep learning options for the encoder $\mathcal{E}$, it can either be a Recurrent Neural Network, or a Convolutional Neural Network. We stress that our proposed similarity measure NeuralWarp is agnostic to the choice of neural architecture and can operate with all encoders that are differentiable with respect to its parameters.

\textit{NeuralWarp}: The proposed similarity function $S: \R^{T \times K} \times \R^{T \times K} \rightarrow \left[0, 1\right] $ is formalized in Equation~\ref{eq:elasticdistance}, which is a generalization of the all-pairs global warping alignment from Equation~\ref{eq:unwarpeddistance}. We propose to measure the warped distance between all pairs of encoded vectors in the latent deep representation $\mathcal{E}$. The distance between the context vectors at the $i$-th index of the first series and the $j$-th index of the second series, i.e. between the vectors $\mathcal{E}(A)_i \in \R^K$ and $\mathcal{E}(B)_j \in \R^K$ is captured by $| \mathcal{E}(A)_i - \mathcal{E}(B)_j |$. This distance is not directly taken into account, but only if a warper function $\Phi\left( \mathcal{E}(A)_i, \mathcal{E}(B)_j \right): \R^K \times \R^K \rightarrow \left[0,1\right]$ decides to align the contexts. We decided to name this similarity model as "NeuralWarp" for paying tribute to the fact that we learn a warping function in a deep series representation.

\begin{eqnarray}
\label{eq:elasticdistance}
\mathcal{S}_{A,B} = \exp\left(- \frac{1}{T^2} \; \sum_{i = 1}^{T} \; \sum_{j = 1}^{T} | \mathcal{E}(A)_i - \mathcal{E}(B)_j | \; \Phi\left( \mathcal{E}(A)_i, \mathcal{E}(B)_j \right) \right) 
\end{eqnarray}

The warper $\Phi: \R^{2K} \rightarrow [0,1] $ is a parametric function that decides smoothly, based on the series contexts $\mathcal{E}(A)_i \in \R^K, i=1,\dots,T$ whether ($\Phi\approx1$) or not ($\Phi\approx0$) to align the two activations at the respective positions $i$ and $j$ of both latent series. Such a function is expressed as a neural network with $\R^{2K}$ inputs and a single output neuron with a Sigmoid activation. The warper is modeled as a fully-connected deep forward network with linear rectifier activation at the neurons of the hidden layers. Finally the warped all-pairs distance is converted to a similarity function by applying the exponential function of the negative distance, since $\exp(-x) \rightarrow 1$ when $x \rightarrow 0$ and  $\exp(-x) \rightarrow 0$ when $x \rightarrow \infty$. We experimented with different p-norms for the context difference $|\mathcal{E}(A)_i - \mathcal{E}(B)_j|$ but found the L1 norm (absolute value) to perform best. It is worth emphasizing that the proposed measure provides a direct probability ($\mathcal{S} \in [0,1]$) of the similarity between series, instead of merely quantifying the inverse of distance.

The proposed deep elastic similarity measure is illustrated in Figure~\ref{fig:mechanism}. Two pairs of similar instances, denoted as $A^{+}, B^{+}$, and dissimilar instances, denoted as $A^{-}, B^{-}$ are inputted to the same encoder network (hence the metaphor Siamese). In the provided illustration, the encoder $\mathcal{E}$ is a bi-directional Recurrent Neural Network. The all-pairs distance matrix and the warping function output are also illustrated as subplots. For this dataset we trained a 5-layers Siamese deep network (1 layer RNN encoder of 16 cells and 4 layers fully-connected warper with [16,8,4,1] neurons). The interpretation of the deep elastic warping is slightly different than the case of raw time series values, since now the warper can alter the deep series representation. In other words, the warper has forced the encoder $\mathcal{E}$ to produce activations in a way that the warper $\Phi$ can achieve small values for similar time series and larger values for dissimilar series. The ultimate aim is to learn $\mathcal{E}, \Phi$ in a way that Equation~\ref{eq:elasticdistance} approaches to one for similar distances and zero for dissimilar ones. As we see from Figure~\ref{fig:mechanism}, the encoder and warper have been trained to produce larger similarities $\mathcal{S}$ for the pair $A^{+}, B^{+}$.

\subsection{Optimization Objective}
\label{sec:optimization}

Having defined the NeuralWarp parametric similarity measure in Section~\ref{sec:deepwarp}, we now derive the loss function for learning the encoder and warper networks' parameters.

We stress out that explicit ground truth similarity annotations for time series data are scarce\footnote{We found no public dataset with an explicit annotation of similar series pairs.}. Yet it is possible to utilize the vast amount of labeled instances for classification purposes by assuming series from the same class are similar. Note that it is an established approach to train and test similarity measures through classification datasets, instead of explicitly annotated pairs of similar series~\cite{Ding:2008:QMT:1454159.1454226,serra_empirical_2014}. In that context, the pairs of similar instances having the same label are defined as $\mathcal{P} = \left\{ (T_n,T_m) \; | \; Y_n = Y_m \right\}$, where $T_n$ denotes the $n$-th series in the training set and $Y_n \in \mathbb{N}$ its corresponding label. In the opposite manner, pairs of instances from different classes are judged to be un-similar as $\mathcal{N} = \left\{ (T_p,T_q) \; | \; Y_p \ne Y_q \right\}$. Overall, the similarity measure is trained by minimizing the logistic loss objective $\mathcal{L}(\mathcal{E}, \Phi)$ of Equation~\ref{eq:optobjective}, which enforces that the similarity measure is much larger for pairs of instances from the same class, compared to pairs of instances belonging to different classes. 

\begin{eqnarray}
\label{eq:optobjective}
\nonumber
\mathcal{L}(\mathcal{E}, \Phi) =: \argmin_{\mathcal{E}, \Phi} \;\;\;  \frac{1}{|\mathcal{P}|} \sum_{(A^{+}, B^{+}) \in \mathcal{P}} \log\left( \mathcal{S}_{A^{+}, B^{+}}(\mathcal{E}, \Phi) \right) \;\;\;\;\;\;\;\; \\ \;\;\;\;\;\;\;  + \;\; \frac{1}{|\mathcal{N}|} \sum_{(A^{-}, B^{-}) \in \mathcal{N}} \log\left(1 - \mathcal{S}_{A^{-}, B^{-}}(\mathcal{E}, \Phi) \right) \;\;\; 
\end{eqnarray}

\subsubsection{Learning Algorithm}

The optimization of the objective function of Equation~\ref{eq:optobjective} is carried out by a Stochastic Gradient Descent learning routine as illustrated in Algorithm~\ref{alg:learning}. In a series of iterations we randomly select a batch of $K$ similar and $K$ dissimilar pairs (line 2). The overall loss of the $i$-th batch of pairs is the aggregation $\mathcal{L}^{(i)}$ of line 3. All the parameters (i.e. neural weights) $\theta_\Phi$ of the warping network and the parameters $\theta_\mathcal{E}$ of the embedding function are updated to minimize the batch loss. In principle, the derivatives $\frac{\partial \mathcal{L}^{(i)}}{\partial \theta}$ are trivially computed by backpropagating the error in minimizing Equation~\ref{eq:optobjective} w.r.t. each of the specific neural weights $\theta$. Nevertheless, we follow the established practice of using the automatic differentiation functionality of deep learning libraries (in our case Tensorflow) for computing $\frac{\partial \mathcal{L}^{(i)}}{\partial \theta}$. In addition, we use the Adam optimizing strategy~\cite{DBLP:journals/corr/KingmaB14} for updating the per-parameter learning rate $\eta^{(i)}_{\theta}$ at every iteration. 

\begin{algorithm}
\SetAlgorithmName{Algorithm}{}{}
\SetKwInOut{Return}{Return}
\SetKwInOut{Input}{Input}
\Input{$\;\;$List of similar series pairs $\mathcal{P}$, List of dissimilar series pairs $\mathcal{N}$, Number of iterations $I$, Batch size $K$, Learning rate $\eta$.}
\begin{doublespace}
	\For{$i=1, \dots, I$}
	{
	$(A^{+}_k, B^{+}_k) \sim \mathcal{P}; \;\;\; (A^{-}_k, B^{-}_k) \sim \mathcal{N}; \;\;\; k=1,\dots,K$\;
	$\mathcal{L}^{(i)} = \frac{1}{K} \sum\limits_{k=1}^{K} \log\left(S_{A^{+}_k, B^{+}_k}\right) + \log\left(1 - S_{A^{-}_k, B^{-}_k}\right)$ \;
	$\theta_\mathcal{E} \leftarrow \theta_\mathcal{E} - \eta^{(i)}_{\theta_\mathcal{E}} \;\; \frac{\partial \mathcal{L}^{(i)}}{\partial \theta_\mathcal{E}} $ \;
	$\theta_\Phi \leftarrow \theta_\Phi - \eta^{(i)}_{\theta_\Phi} \;\; \frac{\partial \mathcal{L}^{(i)}}{\partial \theta_\Phi} $ \;
	}
\end{doublespace}   
\caption{Learning the NeuralWarp similarity}
\Return{$\;\;$Model parameters $\theta_\mathcal{E},\theta_\Phi$}
\label{alg:learning}
\end{algorithm}

\begin{figure}[h!]
	\centering
	\includegraphics[width=0.26\linewidth,trim=3.4cm 5cm 7.9cm 0.5cm, clip=false]{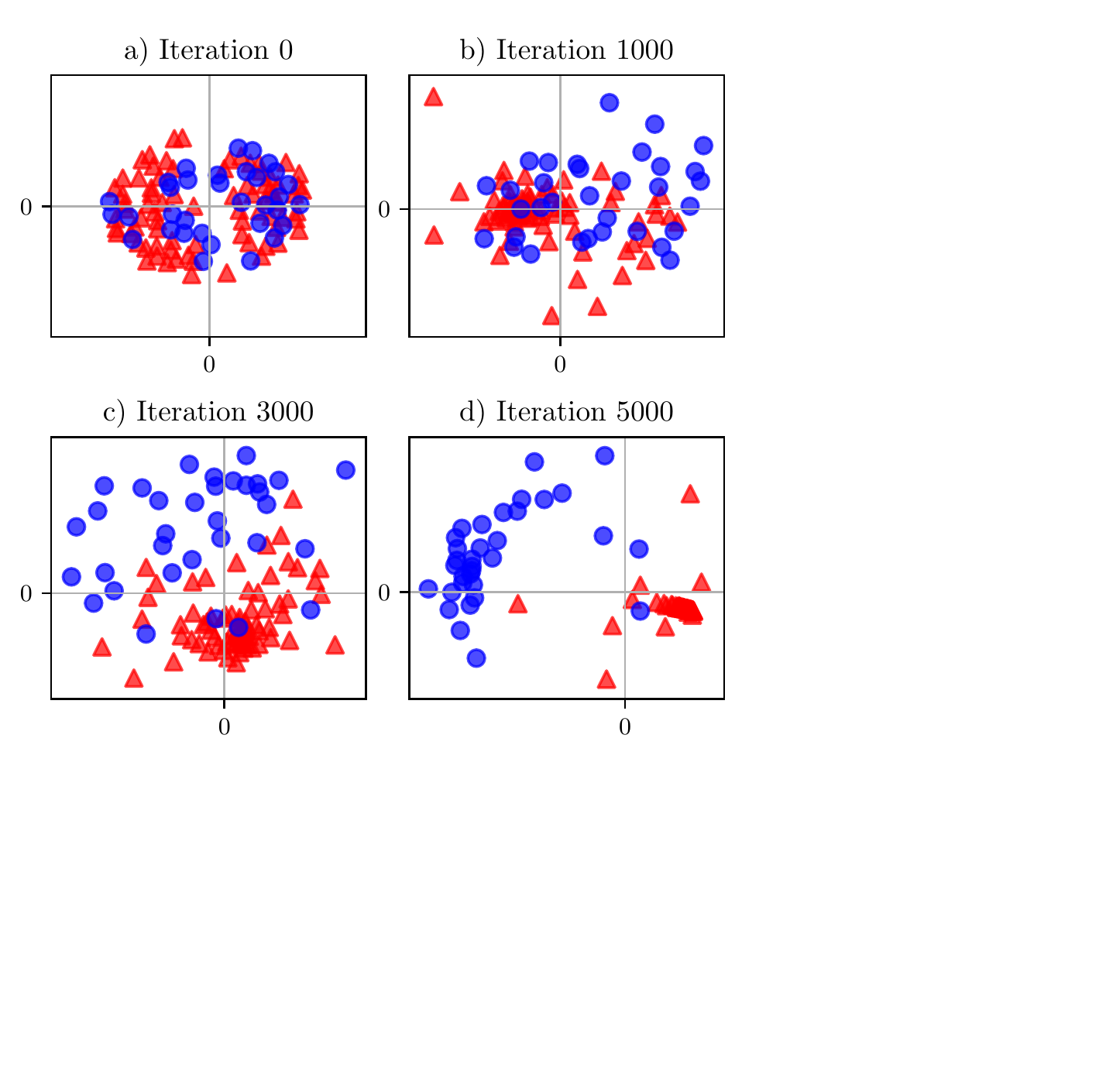}
	\caption{Multidimensional scaling of the pairwise similarities of test series belonging to the two most frequent classes from the HAR dataset~\cite{app7101101}, computed through NeuralWarp with a bi-directional RNN/LSTM encoder.}
	\label{fig:mds}
\end{figure}

We illustrate the progress of a typical execution of the learning algorithm with the assistance of Figure~\ref{fig:mds}. One typical visualization that shows the performance of distance measures is by plotting the distance matrix of a set of test series into a 2-dimensional representation, here computed by the Multidimensional Scaling algorithm. In Figure~\ref{fig:mds} we trained the similarity $\mathcal{S}$ on the HAR dataset~\cite{app7101101} using Algorithm~\ref{alg:learning} and we computed the distance among all test instances for the sake of visualization. Note that our similarity can be trivially converted to a distance (dissimilarity) as $\left(1-\mathcal{S}\right) \in \left[0,1\right]$. The detailed protocol with the settings of the NeuralWarp model is further clarified in Section~\ref{sec:protocol} (i.e. choice of network size, $K$, $\eta$, etc.). Testing series from the two most frequent classes of the HAR dataset~\cite{app7101101} were selected and their distance matrix was reduced to two dimensional representations in the subplots. In the beginning of the optimization procedure (Iteration 0), the distances between similar series are not distinguishable from the dissimilar ones. We notice that after 5000 iterations the intra-class similarities start to become larger than the inter-class similarities, producing a noticeable segregation among the two classes.

\subsection{Connection to Attention Models}

The warped similarity model we propose shares some intuition with the Attention mechanism in language translation~\cite{bahdanau+al-2014-nmt}. Attention, which is typically used for neural translation, models the impact of the words from the sentences of the origin language into predicting the translated words of the target language sentence. In certain aspects, we can think of the contextual function that decides the impact of words in a sentence as some sort of warping function. Therefore, even though attention models solve a prediction task (predict/translate a sequence), we still can categorize our model as a form of generalized attention for time-series similarity. 

\section{Experiments}
\label{sec:empirical}

\subsection{Baselines}
\label{sec:baselines}

The experimental results of this section aim at validating the research hypothesis of the paper specified in Section~\ref{sec:novelty}. There are three competing approaches when referring to similarity measures in time series:

\subsubsection*{\bf Non-parametric:} 
	The first class of similarity measures are the ones that do not have parameters that need to be learned using training data and can be deployed as static measures to any dataset. Among the most prominent and successful measures are the Dynamic Time Warping (DTW) and Time Warped Edit Distance (TWED) \cite{Ding:2008:QMT:1454159.1454226, serra_empirical_2014}.
	
\subsubsection*{\bf Siamese Deep Similarity:} This second class of measures relies on projecting sequential data in a deep representation, where similar series should have small distances. The neural networks of the Siamese deep similarities are either Convolutional Neural Networks (CNN)~\cite{Che2017DECADEA} or Recurrent Neural Networks(RNN)~\cite{mueller_siamese_2016,pei_modeling_2016}. The distance in the deep representation is expressed as the $L_p$ norm (typically L1 and L2) of the latent embedding vectors (i.e. convolutional feature maps, or LSTM cells' activations)~\cite{mueller_siamese_2016, pei_modeling_2016, roy_action_2018}. That simply means $\mathcal{S}_{A,B} = \exp\left(- \frac{1}{T} \; \sum_{i = 1}^{T} | \mathcal{E}(A)_i - \mathcal{E}(B)_i | \right)$.
	
\subsubsection*{\bf NeuralWarp:}

The final type of similarity measures is the NeuralWarp approach that we proposed in Section~\ref{sec:deepwarp}, which extends the Siamese Deep Similarity measures by introducing a parametric warping neural network in the deep representation layer. Similar to the case of the Siamese architectures, the embedding network is either a CNN or RNN. The comparison of NeuralWarp with plain Siamese similarity measures will shed light on the usefulness of having a warping functionality in the deep encoding. On the other hand, comparing NeuralWarp against non-parametric similarity measures will help understanding the benefit of deep learning for time-series similarities.

\begin{table*}[h!]
	\centering 
	\caption{Classification accuracy comparison against similarity measures}
	\label{tab:results}
	\begin{tabular}{lcccccccccc}
		\toprule
		\multicolumn{1}{l}{Dataset} & 
		\multicolumn{4}{c}{Dataset Description} & \multicolumn{2}{c}{Non-parametric} & \multicolumn{2}{c}{Siamese} & \multicolumn{2}{c}{NeuralWarp} \\ \cmidrule(l{5pt}llr{5pt}){2-5} \cmidrule(l{5pt}r{5pt}){6-7} \cmidrule(l{5pt}r{5pt}){8-9} \cmidrule(l{5pt}r{5pt}){10-11}  
		& Train-Test & Classes & Length & Channels & DTW & TWED & CNN & RNN & CNN-W & RNN-W \\
		\cmidrule(l{5pt}r{5pt}){1-1}
		\cmidrule(l{5pt}llr{5pt}){2-5} \cmidrule(l{5pt}r{5pt}){6-7} \cmidrule(l{5pt}r{5pt}){8-9} \cmidrule(l{5pt}r{5pt}){10-11}
		Satellite & 89720-9967 & 9 & 23 & 10 & 0.8781 & 0.7630 & 0.9166 & 0.8867 & \textbf{0.9295} & 0.9206 \\ 
		Speech & 59987-6471 & 30 & 71 & 13 & 0.6525 & 0.0355 & 0.2896 & 0.7894  & 0.8605 & \textbf{0.8969} \\ 
		Crop & 21600-2400 & 24 & 46 & 1 & 0.7553 & \textbf{0.7690} & 0.6433 & 0.7029 & 0.6117 & 0.7604  \\ 
		HAR & 3531-392 & 17 & 57 & 3 & 0.7500 & 0.7678 & 0.9566 & 0.9642 & 0.9617 & \textbf{0.9719} \\ \midrule
		\multicolumn{5}{c}{\emph{Wins}} & 0 & 1 & 0 & 0 & 1 & 2  \\
		\multicolumn{5}{c}{\emph{Ranks}} & 4.5 $\pm$ 1.2 & 4.5 $\pm$ 2.38 & 4.3 $\pm$ 0.9 & 3.2 $\pm$ 0.9 & 3.0 $\pm$ 2.1& 1.5 $\pm$ 0.6 \\ \toprule
	\end{tabular}
\end{table*}

\subsection{Experimental Protocol}
\label{sec:protocol}

Deep Learning requires several hyper-parameters that define the complexity of the neural architectures, as well as the optimization procedure. The optimal approach of setting the hyper-parameters is by tuning them via cross-validation. Unfortunately, given the large computational demands (see run-times in Section~\ref{sec:runtime}) of deep learning architectures, a proper hyper-parameter search is unfeasible. Under these circumstances we follow the established trend of using one specific deep architecture designed based on expert knowledge. However, in order to be fair to the baselines, we used the same CNN and RNN model complexity for both the un-warped Siamese network, as well as the NeuralWarp approach. The CNN architecture we used for the encoder is a three layers deep network with 1024 filters on the first layer, 128 filters on the second and 64 on the third layer. The filter sizes are 5, 5, 3 per each respective layer and the magnitude of applied strides 2, 1, 1. After each convolutional layer we applied a batch normalization operation followed by a ReLU activation function. After the third convolutional layer we applied a Dropout regularization layer with a drop rate of 5\%.  

On the other hand, the RNN encoder is a three-layers deep bi-directional LSTM network with 256, 128 and 64 cells per layer, where the hyperbolic tangent activation was used at each cell. In the end of the LSTM network we apply a batch normalization layer followed again by a 5\% Dropout. The warper network function is the heart of the NeuralWarp model and is composed of a fully-connected neural network architecture with three layers of 64, 16, and 1 neurons each. We use the Linear Rectifier function as the activation of the neurons, but we did not add batch normalization and Dropout since they were already added at the encoder. 

In terms of the hyper-parameters needed for the optimization routine, we trained all models for 1M batches ($I$ at Algorithm~\ref{alg:learning}) with 100 pairs, 50 positive and 50 negative ($K$ at Algorithm~\ref{alg:learning}) drawn randomly per batch. In order to minimize the loss we applied stochastic gradient descent update steps using the Adam optimizer \cite{DBLP:journals/corr/KingmaB14} with an initial learning rate of $10^{-3}$. When facing a divergence during training, i.e. loss increasing instead of decreasing, we restarted the optimization with a smaller learning rate of $10^{-4}$. In fact a divergence happened only once with the warped CNN encoder at the Crop dataset. Finally, in order to avoid exploding gradients we truncated the gradient values to a maximum magnitude of 10. Our implementation of the warped and unwarped deep networks, and the optimization routine, was coded using the Tensorflow library. To promote reproducibility we are making the implementation public\footnote{\url{https://github.com/josifgrabocka/neuralwarp}}.

Regarding the non-parametric baselines, we used the fast DTW implementation of \cite{salvador_toward_2007} through the official python package\footnote{https://pypi.org/project/fastdtw/}. The TWED baseline requires two hyper-parameters, which were set as $1.0$ penalty and $0.001$ stiffness coefficient, following the findings of a survey on the best TWED hyper-parameters~\cite{serra_empirical_2014}.

The evaluation criterion for the performance of similarity measures is the classification accuracy achieved by a Nearest Neighbor classifier that uses the similarity measure on the test split of the datasets, similarly to prior work~\cite{Ding:2008:QMT:1454159.1454226,serra_empirical_2014}. Naturally, for the warped and unwarped Siamese architectures we predict the label of a test series as the label of the training series with the highest similarity score of Equation~\ref{eq:elasticdistance}, while for the non-parametric measures the label of the training instance with the smallest distance.

\subsection{Datasets}

We are going to test the performance of the aforementioned baselines in a set of four real-life time-series datasets. In contrast to the previous research on non-parametric distance measures (e.g. \cite{serra_empirical_2014}) we are not going to base the experiments only on the UCR collection~\cite{UCRArchive2018}. This collection deserves big merits for helping the research community on time series, however it contains a large set of uni-variate, small and simplified datasets~\cite{Hu2013TimeSC} where parametric models tend to over-fit. As a result, it is not a fair and suitable benchmark for large deep learning models. Yet, for the sake of completeness we included the \textit{largest} univariate dataset from the UCR collection named Crop. The selected datasets are:

\begin{itemize}
	\item \textbf{Satellite}-monitored geographical areas through time by measuring the reflectances of satellite images\footnote{Accessed online at  \url{sites.google.com/site/dinoienco/tiselc} on 10.12.2018.}. The task is to classify the image into ten types of landscapes, such as forest, water, etc. 
	\item Google \textbf{Speech}: includes audio recordings for predicting 30 types of words~\cite{speechcommands}. We applied the standard Mel-frequency cepstral coefficients pre-processing technique to convert the raw frequency files into multivariate time series.  	
	\item \textbf{Crop} is another satellite image collection of time series and is the largest univariate dataset of the UCR collection~\cite{UCRArchive2018}. 
	\item \textbf{HAR} represents time series collected via smartphone accelerators~\cite{app7101101}. The problem demands the classification of the type of action that a human subject is performing.	
\end{itemize}

The characteristics of the datasets are detailed in Table~\ref{tab:results}. We have randomly split each dataset into 90\% train and 10\% test (hold-out) disjoint partitions.

\subsection{Results}
\label{sec:results}

The experimental results that aim at comparing NeuralWarp to the prior work on similarity measures are shown in Table~\ref{tab:results}. The displayed values represent the classification rates of a Nearest Neighbor classifier for each respective similarity measure. We have grouped the baselines into the non-parametric measures (DTW, TWED), the Siamese CNN and RNN deep similarity measure, and the proposed NeuralWarp with a CNN and RNN encoder (denoted CNN-W and RNN-W), as clarified in Section~\ref{sec:baselines}.

The results indicate that NeuralWarp is more accurate than the non-parametric measures. Concretely, RNN-W outperforms DTW in all the four datasets, while RNN-W outperforms TWED in three three out of four datasets. On the other hand, CNN-W improves over both DTW and TWED in three out of four datasets. The only dataset where the non-parametric measures outperform any of the NeuralWarp variants is Crop. This dataset appears to have a low signal-to-noise ratio, where parametric models tend to overfit. We found out that TWED achieved a very poor accuracy on the Speech dataset, which could be due to the fact that the recommended hyper-parameters for TWED~\cite{serra_empirical_2014}, detailed in Section~\ref{sec:protocol}, might not be optimal for speech classification. Unfortunately, it took 9 days of time parallelized on 100 CPUs to test TWED on Speech, making it infeasible to re-tune the hyper-parameters in a cross-validation setting. Overall, Table~\ref{tab:results} indicate that NeuralWarp is convincingly more accurate than non-parametric methods.

\begin{figure}[h!]
	\centering
	\includegraphics[scale=0.27,trim={1cm 2cm 0cm 0.8cm}]{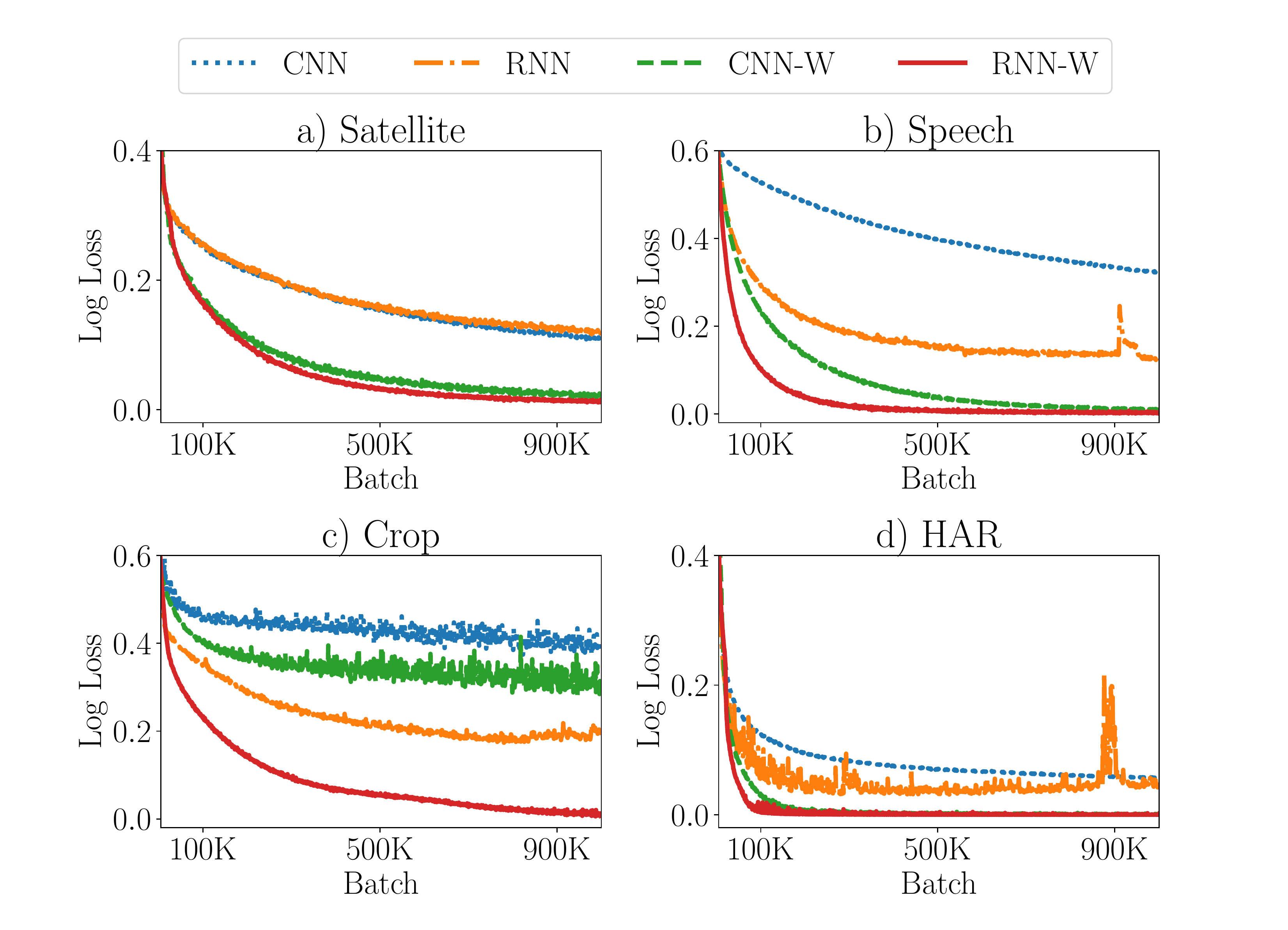}
	\caption{Training set log-Loss fitting of the plain and warped versions of both CNN- and RNN-based encoders}
	\label{fig:fitting}
\end{figure}

The other aspect worth noting is that NeuralWarp is more accurate than the un-warped Siamese approaches. CNN-W is more accurate than CNN in three out of four datasets, while RNN-W outperforms RNN in all the datasets. The empirical results strongly indicate that the warped deep similarity measures perform better than prior work on un-warped Siamese baselines. We noticed that the warped similarity measure fit the loss of Equation~\ref{eq:optobjective} significantly better than the un-warped alternatives as shown in Figure~\ref{fig:fitting}.

The addition of the elastic warping function $\Phi$ makes the NeuralWarp variants CNN-W and RNN-W reduce the similarity log-loss much faster than CNN and RNN in all datasets, as Figure~\ref{fig:fitting} illustrates. The reason for this phenomenon comes from the fact that a warping function helps aligning contexts from $\mathcal{E}$ in cases of intra-class variations such as pattern shifts, local distortions, etc. In contrast to the un-warped deep learning methods, Figure~\ref{fig:fitting} illustrates the advantage of the warping function $\Phi$ in dealing with those intra-class patterns. A devil's advocate might immediately state that the better fitness achieved by the additive model capacity/size comes from the parameters of the warper function. However, this argument is not valid because the warper function is only a tiny fraction of the total parameters of the complete Siamese architecture (see Section~\ref{sec:runtime}). 

The last discussion element those results inspire is: Which encoder produces a better warping, CNN-W or RNN-W? Apparently, encoders based on the LSTM RNN yield more accurate similarity measures, since RNN-W outperforms CNN-W in three datasets. Bringing the arguments of Section~\ref{sec:paramwarpingsim} to attention, we stress out that RNN provides a better context $\mathcal{E}$ for the warping function $\Phi$, because LSTM cells' activations at $\mathcal{E}_i$ accumulate the patterns until index $i$. By contrast, CNN's deep feature map at $\mathcal{E}_i$ represent the values around the $i$-th index of the series, but not the accumulated pattern so far. In that aspect, while CNNs are a successful architecture for detecting patterns through learned filters/kernel, they do not provide accumulated contexts such as the RNN (a.k.a. LSTM states). Therefore, RNN provide better contexts $\mathcal{E}$ than CNN, which is also reflected on the empirical results. To sum up, we conclude that the NeuralWarp with an RNN encoder is the best deep learning similarity measure in terms of accuracy.

\subsubsection{Running Time Overhead}
\label{sec:runtime}

The warping mechanism we propose yields a significant lift with respect to the prior work on similarity measures, in particular against the un-warped Siamese neural network architectures. However, one should ask whether the \textit{"no free lunch"} theorem applies when comparing other quality criteria of the warped vs. un-warped deep models, for instance running times. Table~\ref{tab:train_test_times} shows the training and testing run times for the deep learning approaches in terms of GPU hours. The experiments were carried out in diverse environments: The methods on the Satellite and Speech datasets were trained and tested using Nvidia GeForce GTX 1080 Ti GPU servers, while we purchased additional computational power to run the methods on Crop and HAR from the Google Cloud platform, by renting Nvidia Tesla V100 GPU machines. All-inclusive the experiments of Table~\ref{tab:results} took 725 hours of GPU training time and 577 hours of GPU testing time, i.e. 54.25 GPU days. 

\begin{table}[h!]
	\centering 
	\caption{Training/Testing times in GPU hours}
	\label{tab:train_test_times}
	\begin{tabular}{lcccc}
		\toprule
		\multicolumn{1}{l}{Dataset} & 
		\multicolumn{2}{c}{Siamese} & \multicolumn{2}{c}{NeuralWarp} \\ \cmidrule(l{5pt}llr{5pt}){2-3}
		\cmidrule(l{5pt}llr{5pt}){4-5}
		& CNN & RNN & CNN-W & RNN-W \\
		\cmidrule(l{5pt}r{5pt}){1-1}
		\cmidrule(l{5pt}llr{5pt}){2-3}
		\cmidrule(l{5pt}llr{5pt}){4-5}		
		Satellite & 16/30 & 25/80 & 32/69 & 47/113 \\ 
		Speech & 26/28 & 78/39 & 47/32 & 95/92 \\ 
		Crop & 24/23 & 56/22 & 38/23 & 65/24 \\ 
		SHAR & 21/0.5 & 41/0.6 & 36/0.7 & 78/0.8 \\ \midrule
		Number of weights & 1.41M & 1.04M & 1.42M & 1.05M \\
		\toprule
	\end{tabular}
\end{table}

One notices that the warped versions of the Siamese deep architectures are less than twice slower compared to the un-warped variants. Indeed, the \textit{"no free lunch"} theorem seems to perfectly apply in our case. Nevertheless, a runtime overhead of a fractional proportion is not a showstopper in typical real-life domains. We additionally presented the total number of parameters that each architecture has. All models have more than one million parameters, which can be considered a \textit{'tiny'} architecture by state-of-the-art deep learning examples in other domains, such image recognition. Yet we reason that the sizes of our datasets are not in the range of millions of instances, but only in dozens of thousands.  

The warping function introduces a small overhead in terms of additive parameters (ca. $1\%$). Nonetheless, the runtime overhead that it creates is considerably larger than its size overhead. Such an effect comes from the nature of the back-propogation algorithm. For instance, the gradient of the similarity measure of Equation~\ref{eq:elasticdistance} w.r.t. the encoding of the $i$-th index of the first series $\mathcal{E}(A)$:

\begin{eqnarray}
\label{eq:gradwarp}
\frac{\partial \mathcal{S}_{A,B}}{\partial \mathcal{E}(A)_i} = - \frac{\mathcal{S}_{A,B}}{T^2}  \;  \sum_{j = 1}^{T} \frac{\mathcal{E}(A)_i - \mathcal{E}(B)_j}{| \mathcal{E}(A)_i - \mathcal{E}(B)_j |} \; \frac{\partial \Phi\left( \mathcal{E}(A)_i, \mathcal{E}(B)_j \right) }{\partial \mathcal{E}(A)_i} 
\end{eqnarray}

As we can see from Equation~\ref{eq:gradwarp}, the back-propagation algorithm needs to compute the derivative $\frac{\partial \Phi}{\partial \mathcal{E}}$ of the warping function with respect to all encoder indices of all pairs. For this reason, a small $1\%$ addition in the number of parameters of $\Phi$ might create up to $200\%$ runtime overhead.

\subsubsection{Comparison to Classification Models}

Since our experiments are based on classification datasets, a skeptic reader can raise concerns regarding the need of training similarity measures instead of directly using classification models. First of all, it is important to note that similarity learning is solving a different problem than instance-wise classification, because the problem of similarity is defined in pairs of instances.

\begin{figure}[h!]
	\centering
	\includegraphics[width=0.57\linewidth,trim=5cm 6.4cm 4.5cm 0.3cm, clip=false]{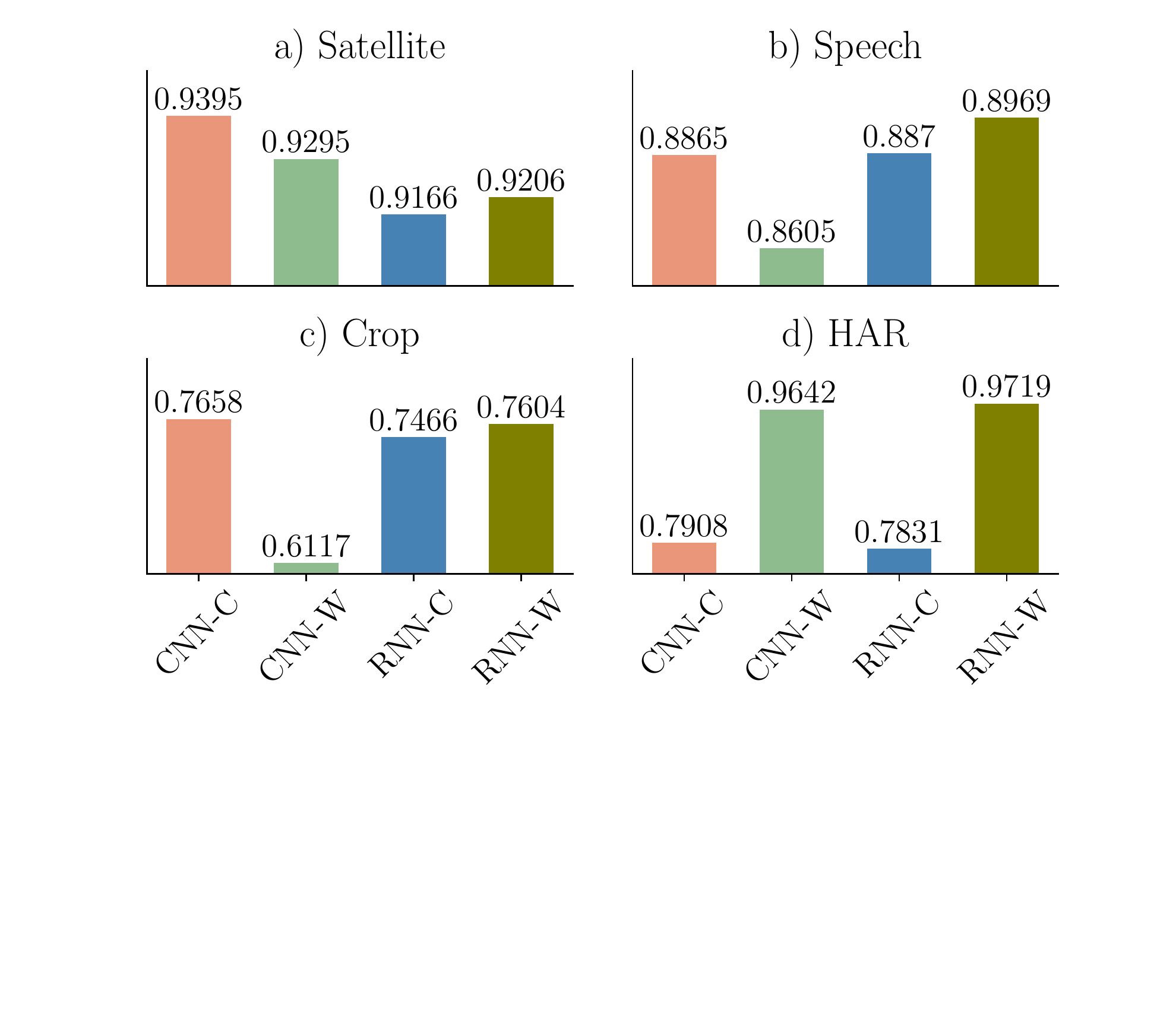}
	\caption{Comparison against deep networks trained for classification (\textit{X}NN-C: \textit{X}NN trained for classification, \textit{X}NN-W: \textit{X}NN trained for similarity; \textit{X} $\in$ \{C, R\})}
	\label{fig:comparisonclassification}
\end{figure}

Furthermore, classification models cannot operate in cases when a series of a particular class occurs in the test set, but not in the training set, while similarity measures can still output the similarity score of any pairs of series. Yet, from a methodological perspective it is interesting to compare the performance of NeuralWarp against neural networks trained for time-series classification. 

In order to compare the classification vs. similarity approaches with respect to classification accuracy we used the same architecture for both types of models. In other words, we used the same CNN and RNN architecture as the ones of the encoders explained in Section~\ref{sec:protocol}. However, the encoded representations were aggregated to a mean vector as $Z_A = \frac{1}{T} \sum_{i=1}^{T} \mathcal{E}(A)_i$ for all series $A$, instead of using a fully connected layer. It has been recently shown that a global aggregation, such as the mean value, improves the performance of deep learning models for time series classification~\cite{wang_time_2017}. Afterwards, we connected the aggregated vector $Z_A$ to a softmax layer for predicting the class of series $A$. 

The results of Figure~\ref{fig:comparisonclassification} show that the classification-trained CNN outperforms the similarity-trained CNN in three out of four datasets. That is not a surprising finding, since a classification-trained model is expected to perform better than a similarity-trained one, because constructing decision boundaries between classes is an easier task than learning a latent representation where all instances of a class are located close to one-another. To help with the intuition, imagine two instances of two different classes that are located nearby, yet on opposite sides of a decision boundary. These two points have a closer distance between them, than to other instances of their respective classes. This behavior is not a problem for a classification model, as long as the decision boundary can split the classes, but is a drawback for similarity measures where proximity matters. However, quite on the contrary, the warped RNN model (RNN-W) is better than the classification RNN on all the datasets. In the light of the results of Figure~\ref{fig:comparisonclassification}, the fact that warped RNN outperformed the classification RNN shows the additive power of the warping function, which makes warped similarities a candidate for classification tasks, too.

\section{Discussion}

Having an accurate deep similarity measure can be instrumental in uniting the vast research in deep learning with the time-series community. The potential for expanding the deep similarity measures spans diverse application domains, such as clustering where NeuralWarp can be used to optimize the cluster centroids directly. 

It is true that Deep Learning can be a painful experience in case where the computational resources in terms of GPUs are scarce, however it seems to be a gentle obstacle considering the widespread application of Deep Learning. In addition, given the evidence on the superior performance of deep neural networks, we forecast a steady rise in the usage of deep models for time-series mining. 

\section{Conclusion}

Similarity measures are at the core of many time-series mining problems and represent an important task for the machine learning community. Unfortunately, deep learning models are not thoroughly explored for time series similarity and the existing Siamese models do not capture the intra-class variations of time series. 

In this paper we propose to learn a warping function for aligning the indices of time series in a deep latent representation. We compared the suggested architecture with two types of encoders (CNN, or RNN) and a deep forward network as a warping function. Experimental comparisons to non-parametric and un-warped Siamese networks demonstrated that the proposed elastic deep similarity measure is more accurate than prior models.

\section*{Acknowledgement}

We acknowledge the funding provided by the "Zentrales Innovationsprogramm Mittelstand" of the German Federal Ministry for Economic Affairs and Energy through the project ADDA. The authors thank Jonas Falkner (Uni. Hildesheim) for the valuable advice.

\bibliographystyle{ACM-Reference-Format}
\bibliography{deepwarping}


\begin{thebibliography}{41}


\ifx \showCODEN    \undefined \def \showCODEN     #1{\unskip}     \fi
\ifx \showDOI      \undefined \def \showDOI       #1{#1}\fi
\ifx \showISBNx    \undefined \def \showISBNx     #1{\unskip}     \fi
\ifx \showISBNxiii \undefined \def \showISBNxiii  #1{\unskip}     \fi
\ifx \showISSN     \undefined \def \showISSN      #1{\unskip}     \fi
\ifx \showLCCN     \undefined \def \showLCCN      #1{\unskip}     \fi
\ifx \shownote     \undefined \def \shownote      #1{#1}          \fi
\ifx \showarticletitle \undefined \def \showarticletitle #1{#1}   \fi
\ifx \showURL      \undefined \def \showURL       {\relax}        \fi
\providecommand\bibfield[2]{#2}
\providecommand\bibinfo[2]{#2}
\providecommand\natexlab[1]{#1}
\providecommand\showeprint[2][]{arXiv:#2}

\bibitem[\protect\citeauthoryear{Bagnall, Lines, Bostrom, Large, and
  Keogh}{Bagnall et~al\mbox{.}}{2017}]%
        {Bagnall2017}
\bibfield{author}{\bibinfo{person}{Anthony Bagnall}, \bibinfo{person}{Jason
  Lines}, \bibinfo{person}{Aaron Bostrom}, \bibinfo{person}{James Large}, {and}
  \bibinfo{person}{Eamonn Keogh}.} \bibinfo{year}{2017}\natexlab{}.
\newblock \showarticletitle{The great time series classification bake off: a
  review and experimental evaluation of recent algorithmic advances}.
\newblock \bibinfo{journal}{\emph{Data Mining and Knowledge Discovery}}
  \bibinfo{volume}{31}, \bibinfo{number}{3} (\bibinfo{date}{01 May}
  \bibinfo{year}{2017}), \bibinfo{pages}{606--660}.
\newblock
\showISSN{1573-756X}
\urldef\tempurl%
\url{https://doi.org/10.1007/s10618-016-0483-9}
\showDOI{\tempurl}


\bibitem[\protect\citeauthoryear{Bahdanau, Cho, and Bengio}{Bahdanau
  et~al\mbox{.}}{2014}]%
        {bahdanau+al-2014-nmt}
\bibfield{author}{\bibinfo{person}{Dzmitry Bahdanau},
  \bibinfo{person}{Kyunghyun Cho}, {and} \bibinfo{person}{Yoshua Bengio}.}
  \bibinfo{year}{2014}\natexlab{}.
\newblock \showarticletitle{Neural Machine Translation by Jointly Learning to
  Align and Translate}.
\newblock \bibinfo{journal}{\emph{arXiv e-prints}}
  \bibinfo{volume}{abs/1409.0473} (\bibinfo{date}{Sept.} \bibinfo{year}{2014}).
\newblock
\urldef\tempurl%
\url{https://arxiv.org/abs/1409.0473}
\showURL{%
\tempurl}


\bibitem[\protect\citeauthoryear{Bromley, Guyon, LeCun, S\"{a}ckinger, and
  Shah}{Bromley et~al\mbox{.}}{1993}]%
        {Bromley:1993:SVU:2987189.2987282}
\bibfield{author}{\bibinfo{person}{Jane Bromley}, \bibinfo{person}{Isabelle
  Guyon}, \bibinfo{person}{Yann LeCun}, \bibinfo{person}{Eduard S\"{a}ckinger},
  {and} \bibinfo{person}{Roopak Shah}.} \bibinfo{year}{1993}\natexlab{}.
\newblock \showarticletitle{Signature Verification Using a "Siamese" Time Delay
  Neural Network}. In \bibinfo{booktitle}{\emph{Proceedings of the 6th
  International Conference on Neural Information Processing Systems}}
  \emph{(\bibinfo{series}{NIPS'93})}. \bibinfo{publisher}{Morgan Kaufmann
  Publishers Inc.}, \bibinfo{address}{San Francisco, CA, USA},
  \bibinfo{pages}{737--744}.
\newblock
\urldef\tempurl%
\url{http://dl.acm.org/citation.cfm?id=2987189.2987282}
\showURL{%
\tempurl}


\bibitem[\protect\citeauthoryear{Bromley, Guyon, LeCun, S{\"a}ckinger, and
  Shah}{Bromley et~al\mbox{.}}{1994}]%
        {bromley1994signature}
\bibfield{author}{\bibinfo{person}{Jane Bromley}, \bibinfo{person}{Isabelle
  Guyon}, \bibinfo{person}{Yann LeCun}, \bibinfo{person}{Eduard S{\"a}ckinger},
  {and} \bibinfo{person}{Roopak Shah}.} \bibinfo{year}{1994}\natexlab{}.
\newblock \showarticletitle{Signature verification using a" siamese" time delay
  neural network}. In \bibinfo{booktitle}{\emph{Advances in neural information
  processing systems}}. \bibinfo{pages}{737--744}.
\newblock


\bibitem[\protect\citeauthoryear{Che}{Che}{2017}]%
        {Che2017DECADEA}
\bibfield{author}{\bibinfo{person}{Zhengping Che}.}
  \bibinfo{year}{2017}\natexlab{}.
\newblock \showarticletitle{DECADE : A Deep Metric Learning Model for
  Multivariate Time Series}. In \bibinfo{booktitle}{\emph{MiLeTs 2017, 3th
  SIGKDD Workshop on Mining and Learning from Time Series}}.
\newblock


\bibitem[\protect\citeauthoryear{Chen and Ng}{Chen and Ng}{2004}]%
        {Chen:2004:MLE:1316689.1316758}
\bibfield{author}{\bibinfo{person}{Lei Chen} {and} \bibinfo{person}{Raymond
  Ng}.} \bibinfo{year}{2004}\natexlab{}.
\newblock \showarticletitle{On the Marriage of Lp-norms and Edit Distance}. In
  \bibinfo{booktitle}{\emph{Proceedings of the Thirtieth International
  Conference on Very Large Data Bases - Volume 30}}
  \emph{(\bibinfo{series}{VLDB '04})}. \bibinfo{publisher}{VLDB Endowment},
  \bibinfo{pages}{792--803}.
\newblock
\showISBNx{0-12-088469-0}
\urldef\tempurl%
\url{http://dl.acm.org/citation.cfm?id=1316689.1316758}
\showURL{%
\tempurl}


\bibitem[\protect\citeauthoryear{Cuturi}{Cuturi}{2011}]%
        {cuturi_fast_2011}
\bibfield{author}{\bibinfo{person}{Marco Cuturi}.}
  \bibinfo{year}{2011}\natexlab{}.
\newblock \showarticletitle{Fast {Global} {Alignment} {Kernels}}. In
  \bibinfo{booktitle}{\emph{Proceedings of the 28th {International}
  {Conference} on {International} {Conference} on {Machine} {Learning}}}
  \emph{(\bibinfo{series}{{ICML}'11})}. \bibinfo{publisher}{Omnipress},
  \bibinfo{address}{USA}, \bibinfo{pages}{929--936}.
\newblock
\showISBNx{978-1-4503-0619-5}
\urldef\tempurl%
\url{http://dl.acm.org/citation.cfm?id=3104482.3104599}
\showURL{%
\tempurl}


\bibitem[\protect\citeauthoryear{Cuturi and Blondel}{Cuturi and
  Blondel}{2017}]%
        {cuturi_soft-dtw:_2017}
\bibfield{author}{\bibinfo{person}{Marco Cuturi} {and} \bibinfo{person}{Mathieu
  Blondel}.} \bibinfo{year}{2017}\natexlab{}.
\newblock \showarticletitle{Soft-{DTW}: a {Differentiable} {Loss} {Function}
  for {Time}-{Series}}. In \bibinfo{booktitle}{\emph{International {Conference}
  on {Machine} {Learning}}}. \bibinfo{pages}{894--903}.
\newblock
\urldef\tempurl%
\url{http://proceedings.mlr.press/v70/cuturi17a.html}
\showURL{%
\tempurl}


\bibitem[\protect\citeauthoryear{Dau, Keogh, Kamgar, Yeh, Zhu, Gharghabi,
  Ratanamahatana, Yanping, Hu, Begum, Bagnall, Mueen, and Batista}{Dau
  et~al\mbox{.}}{2018}]%
        {UCRArchive2018}
\bibfield{author}{\bibinfo{person}{Hoang~Anh Dau}, \bibinfo{person}{Eamonn
  Keogh}, \bibinfo{person}{Kaveh Kamgar}, \bibinfo{person}{Chin-Chia~Michael
  Yeh}, \bibinfo{person}{Yan Zhu}, \bibinfo{person}{Shaghayegh Gharghabi},
  \bibinfo{person}{Chotirat~Ann Ratanamahatana}, \bibinfo{person}{Yanping},
  \bibinfo{person}{Bing Hu}, \bibinfo{person}{Nurjahan Begum},
  \bibinfo{person}{Anthony Bagnall}, \bibinfo{person}{Abdullah Mueen}, {and}
  \bibinfo{person}{Gustavo Batista}.} \bibinfo{year}{2018}\natexlab{}.
\newblock \bibinfo{title}{The UCR Time Series Classification Archive}.
\newblock
\newblock
\newblock
\shownote{\url{https://www.cs.ucr.edu/~eamonn/time_series_data_2018/}.}


\bibitem[\protect\citeauthoryear{Ding, Trajcevski, Scheuermann, Wang, and
  Keogh}{Ding et~al\mbox{.}}{2008}]%
        {Ding:2008:QMT:1454159.1454226}
\bibfield{author}{\bibinfo{person}{Hui Ding}, \bibinfo{person}{Goce
  Trajcevski}, \bibinfo{person}{Peter Scheuermann}, \bibinfo{person}{Xiaoyue
  Wang}, {and} \bibinfo{person}{Eamonn Keogh}.}
  \bibinfo{year}{2008}\natexlab{}.
\newblock \showarticletitle{Querying and Mining of Time Series Data:
  Experimental Comparison of Representations and Distance Measures}.
\newblock \bibinfo{journal}{\emph{Proc. VLDB Endow.}} \bibinfo{volume}{1},
  \bibinfo{number}{2} (\bibinfo{date}{Aug.} \bibinfo{year}{2008}),
  \bibinfo{pages}{1542--1552}.
\newblock
\showISSN{2150-8097}
\urldef\tempurl%
\url{https://doi.org/10.14778/1454159.1454226}
\showDOI{\tempurl}


\bibitem[\protect\citeauthoryear{Garreau, Lajugie, Arlot, and Bach}{Garreau
  et~al\mbox{.}}{2014}]%
        {garreau_metric_2014}
\bibfield{author}{\bibinfo{person}{Damien Garreau}, \bibinfo{person}{Rémi
  Lajugie}, \bibinfo{person}{Sylvain Arlot}, {and} \bibinfo{person}{Francis
  Bach}.} \bibinfo{year}{2014}\natexlab{}.
\newblock \showarticletitle{Metric {Learning} for {Temporal} {Sequence}
  {Alignment}}.
\newblock In \bibinfo{booktitle}{\emph{Advances in {Neural} {Information}
  {Processing} {Systems} 27}},
  \bibfield{editor}{\bibinfo{person}{Z.~Ghahramani},
  \bibinfo{person}{M.~Welling}, \bibinfo{person}{C.~Cortes},
  \bibinfo{person}{N.~D. Lawrence}, {and} \bibinfo{person}{K.~Q. Weinberger}}
  (Eds.). \bibinfo{publisher}{Curran Associates, Inc.},
  \bibinfo{pages}{1817--1825}.
\newblock
\urldef\tempurl%
\url{http://papers.nips.cc/paper/5383-metric-learning-for-temporal-sequence-alignment.pdf}
\showURL{%
\tempurl}


\bibitem[\protect\citeauthoryear{Gharghabi, Imani, Bagnall, Darvishzadeh, and
  Keogh}{Gharghabi et~al\mbox{.}}{2018}]%
        {mpdist}
\bibfield{author}{\bibinfo{person}{S. Gharghabi}, \bibinfo{person}{S. Imani},
  \bibinfo{person}{A. Bagnall}, \bibinfo{person}{A. Darvishzadeh}, {and}
  \bibinfo{person}{E. Keogh}.} \bibinfo{year}{2018}\natexlab{}.
\newblock \showarticletitle{MPdist: A Novel Time Series Distance Measure to
  Allow Data Mining in More Challenging Scenarios}. In
  \bibinfo{booktitle}{\emph{2018 IEEE 18th International Conference on Data
  Mining (ICDM)}}.
\newblock


\bibitem[\protect\citeauthoryear{Grabocka, Schilling, Wistuba, and
  Schmidt-Thieme}{Grabocka et~al\mbox{.}}{2014}]%
        {Grabocka:2014:LTS:2623330.2623613}
\bibfield{author}{\bibinfo{person}{Josif Grabocka}, \bibinfo{person}{Nicolas
  Schilling}, \bibinfo{person}{Martin Wistuba}, {and} \bibinfo{person}{Lars
  Schmidt-Thieme}.} \bibinfo{year}{2014}\natexlab{}.
\newblock \showarticletitle{Learning Time-series Shapelets}. In
  \bibinfo{booktitle}{\emph{Proceedings of the 20th ACM SIGKDD International
  Conference on Knowledge Discovery and Data Mining}}
  \emph{(\bibinfo{series}{KDD '14})}. \bibinfo{publisher}{ACM},
  \bibinfo{address}{New York, NY, USA}, \bibinfo{pages}{392--401}.
\newblock
\showISBNx{978-1-4503-2956-9}
\urldef\tempurl%
\url{https://doi.org/10.1145/2623330.2623613}
\showDOI{\tempurl}


\bibitem[\protect\citeauthoryear{Hu, Chen, and Keogh}{Hu et~al\mbox{.}}{2013}]%
        {Hu2013TimeSC}
\bibfield{author}{\bibinfo{person}{Bing Hu}, \bibinfo{person}{Yanping Chen},
  {and} \bibinfo{person}{Eamonn~J. Keogh}.} \bibinfo{year}{2013}\natexlab{}.
\newblock \showarticletitle{Time Series Classification under More Realistic
  Assumptions}. In \bibinfo{booktitle}{\emph{Siam International Conference on
  Data Mining (SDM)}}.
\newblock


\bibitem[\protect\citeauthoryear{Hu, Lu, and Tan}{Hu et~al\mbox{.}}{2014}]%
        {hu_discriminative_2014}
\bibfield{author}{\bibinfo{person}{Junlin Hu}, \bibinfo{person}{Jiwen Lu},
  {and} \bibinfo{person}{Yap-Peng Tan}.} \bibinfo{year}{2014}\natexlab{}.
\newblock \showarticletitle{Discriminative {Deep} {Metric} {Learning} for
  {Face} {Verification} in the {Wild}}. \bibinfo{pages}{1875--1882}.
\newblock
\urldef\tempurl%
\url{https://www.cv-foundation.org/openaccess/content_cvpr_2014/html/Hu_Discriminative_Deep_Metric_2014_CVPR_paper.html}
\showURL{%
\tempurl}


\bibitem[\protect\citeauthoryear{Hunt and Szymanski}{Hunt and
  Szymanski}{1977}]%
        {Hunt:1977:FAC:359581.359603}
\bibfield{author}{\bibinfo{person}{James~W. Hunt} {and}
  \bibinfo{person}{Thomas~G. Szymanski}.} \bibinfo{year}{1977}\natexlab{}.
\newblock \showarticletitle{A Fast Algorithm for Computing Longest Common
  Subsequences}.
\newblock \bibinfo{journal}{\emph{Commun. ACM}} \bibinfo{volume}{20},
  \bibinfo{number}{5} (\bibinfo{date}{May} \bibinfo{year}{1977}),
  \bibinfo{pages}{350--353}.
\newblock
\showISSN{0001-0782}
\urldef\tempurl%
\url{https://doi.org/10.1145/359581.359603}
\showDOI{\tempurl}


\bibitem[\protect\citeauthoryear{Karim, Majumdar, Darabi, and Chen}{Karim
  et~al\mbox{.}}{2017}]%
        {fazle_lstm_2017}
\bibfield{author}{\bibinfo{person}{Fazle Karim}, \bibinfo{person}{Somshubra
  Majumdar}, \bibinfo{person}{Houshang Darabi}, {and} \bibinfo{person}{Shun
  Chen}.} \bibinfo{year}{2017}\natexlab{}.
\newblock \showarticletitle{LSTM Fully Convolutional Networks for Time Series
  Classification}.
\newblock \bibinfo{journal}{\emph{IEEE Access}} (\bibinfo{date}{2 12}
  \bibinfo{year}{2017}).
\newblock
\showISSN{2169-3536}
\urldef\tempurl%
\url{https://doi.org/10.1109/ACCESS.2017.2779939}
\showDOI{\tempurl}


\bibitem[\protect\citeauthoryear{Kingma and Ba}{Kingma and Ba}{2014}]%
        {DBLP:journals/corr/KingmaB14}
\bibfield{author}{\bibinfo{person}{Diederik~P. Kingma} {and}
  \bibinfo{person}{Jimmy Ba}.} \bibinfo{year}{2014}\natexlab{}.
\newblock \showarticletitle{Adam: {A} Method for Stochastic Optimization}.
\newblock \bibinfo{journal}{\emph{CoRR}}  \bibinfo{volume}{abs/1412.6980}
  (\bibinfo{year}{2014}).
\newblock
\showeprint[arxiv]{1412.6980}
\urldef\tempurl%
\url{http://arxiv.org/abs/1412.6980}
\showURL{%
\tempurl}


\bibitem[\protect\citeauthoryear{Li and Ogihara}{Li and Ogihara}{2004}]%
        {li_content_2004}
\bibfield{author}{\bibinfo{person}{Tao Li} {and} \bibinfo{person}{M. Ogihara}.}
  \bibinfo{year}{2004}\natexlab{}.
\newblock \showarticletitle{Content-based music similarity search and emotion
  detection}. In \bibinfo{booktitle}{\emph{2004 IEEE International Conference
  on Acoustics, Speech, and Signal Processing(ICASSP)}},
  Vol.~\bibinfo{volume}{05}. \bibinfo{pages}{V--705--8 vol.5}.
\newblock
\showISSN{1520-6149}
\urldef\tempurl%
\url{https://doi.org/10.1109/ICASSP.2004.1327208}
\showDOI{\tempurl}


\bibitem[\protect\citeauthoryear{Marteau}{Marteau}{2009}]%
        {marteau_time_2009}
\bibfield{author}{\bibinfo{person}{Pierre-Francois Marteau}.}
  \bibinfo{year}{2009}\natexlab{}.
\newblock \showarticletitle{Time {Warp} {Edit} {Distance} with {Stiffness}
  {Adjustment} for {Time} {Series} {Matching}}.
\newblock \bibinfo{journal}{\emph{IEEE Trans. Pattern Anal. Mach. Intell.}}
  \bibinfo{volume}{31}, \bibinfo{number}{2} (\bibinfo{date}{Feb.}
  \bibinfo{year}{2009}), \bibinfo{pages}{306--318}.
\newblock
\showISSN{0162-8828}
\urldef\tempurl%
\url{https://doi.org/10.1109/TPAMI.2008.76}
\showDOI{\tempurl}


\bibitem[\protect\citeauthoryear{McFee, Barrington, and Lanckriet}{McFee
  et~al\mbox{.}}{2012}]%
        {mcfee_learning_2012}
\bibfield{author}{\bibinfo{person}{B. McFee}, \bibinfo{person}{L. Barrington},
  {and} \bibinfo{person}{G. Lanckriet}.} \bibinfo{year}{2012}\natexlab{}.
\newblock \showarticletitle{Learning Content Similarity for Music
  Recommendation}.
\newblock \bibinfo{journal}{\emph{IEEE Transactions on Audio, Speech, and
  Language Processing}} \bibinfo{volume}{20}, \bibinfo{number}{8}
  (\bibinfo{date}{Oct} \bibinfo{year}{2012}), \bibinfo{pages}{2207--2218}.
\newblock
\showISSN{1558-7916}
\urldef\tempurl%
\url{https://doi.org/10.1109/TASL.2012.2199109}
\showDOI{\tempurl}


\bibitem[\protect\citeauthoryear{Micucci, Mobilio, and Napoletano}{Micucci
  et~al\mbox{.}}{2017}]%
        {app7101101}
\bibfield{author}{\bibinfo{person}{Daniela Micucci}, \bibinfo{person}{Marco
  Mobilio}, {and} \bibinfo{person}{Paolo Napoletano}.}
  \bibinfo{year}{2017}\natexlab{}.
\newblock \showarticletitle{UniMiB SHAR: A Dataset for Human Activity
  Recognition Using Acceleration Data from Smartphones}.
\newblock \bibinfo{journal}{\emph{Applied Sciences}} \bibinfo{volume}{7},
  \bibinfo{number}{10} (\bibinfo{year}{2017}).
\newblock
\showISSN{2076-3417}
\urldef\tempurl%
\url{https://doi.org/10.3390/app7101101}
\showDOI{\tempurl}


\bibitem[\protect\citeauthoryear{Mueller and Thyagarajan}{Mueller and
  Thyagarajan}{2016}]%
        {mueller_siamese_2016}
\bibfield{author}{\bibinfo{person}{Jonas Mueller} {and} \bibinfo{person}{Aditya
  Thyagarajan}.} \bibinfo{year}{2016}\natexlab{}.
\newblock \showarticletitle{Siamese {Recurrent} {Architectures} for {Learning}
  {Sentence} {Similarity}}. In \bibinfo{booktitle}{\emph{Proceedings of the
  {Thirtieth} {AAAI} {Conference} on {Artificial} {Intelligence}}}
  \emph{(\bibinfo{series}{{AAAI}'16})}. \bibinfo{publisher}{AAAI Press},
  \bibinfo{address}{Phoenix, Arizona}, \bibinfo{pages}{2786--2792}.
\newblock
\urldef\tempurl%
\url{http://dl.acm.org/citation.cfm?id=3016100.3016291}
\showURL{%
\tempurl}


\bibitem[\protect\citeauthoryear{Pei, Tax, and van~der Maaten}{Pei
  et~al\mbox{.}}{2016}]%
        {pei_modeling_2016}
\bibfield{author}{\bibinfo{person}{Wenjie Pei}, \bibinfo{person}{David M.~J.
  Tax}, {and} \bibinfo{person}{Laurens van~der Maaten}.}
  \bibinfo{year}{2016}\natexlab{}.
\newblock \showarticletitle{Modeling {Time} {Series} {Similarity} with
  {Siamese} {Recurrent} {Networks}}.
\newblock \bibinfo{journal}{\emph{arXiv:1603.04713 [cs]}}
  (\bibinfo{date}{March} \bibinfo{year}{2016}).
\newblock
\urldef\tempurl%
\url{http://arxiv.org/abs/1603.04713}
\showURL{%
\tempurl}
\newblock
\shownote{arXiv: 1603.04713.}


\bibitem[\protect\citeauthoryear{Rakthanmanon, Campana, Mueen, Batista,
  Westover, Zhu, Zakaria, and Keogh}{Rakthanmanon et~al\mbox{.}}{2012}]%
        {Rakthanmanon:2012:SMT:2339530.2339576}
\bibfield{author}{\bibinfo{person}{Thanawin Rakthanmanon},
  \bibinfo{person}{Bilson Campana}, \bibinfo{person}{Abdullah Mueen},
  \bibinfo{person}{Gustavo Batista}, \bibinfo{person}{Brandon Westover},
  \bibinfo{person}{Qiang Zhu}, \bibinfo{person}{Jesin Zakaria}, {and}
  \bibinfo{person}{Eamonn Keogh}.} \bibinfo{year}{2012}\natexlab{}.
\newblock \showarticletitle{Searching and Mining Trillions of Time Series
  Subsequences Under Dynamic Time Warping}. In
  \bibinfo{booktitle}{\emph{Proceedings of the 18th ACM SIGKDD International
  Conference on Knowledge Discovery and Data Mining}}
  \emph{(\bibinfo{series}{KDD '12})}. \bibinfo{publisher}{ACM},
  \bibinfo{address}{New York, NY, USA}, \bibinfo{pages}{262--270}.
\newblock
\showISBNx{978-1-4503-1462-6}
\urldef\tempurl%
\url{https://doi.org/10.1145/2339530.2339576}
\showDOI{\tempurl}


\bibitem[\protect\citeauthoryear{Roy, Mohan, and Murty}{Roy
  et~al\mbox{.}}{2018}]%
        {roy_action_2018}
\bibfield{author}{\bibinfo{person}{D. Roy}, \bibinfo{person}{C.~K. Mohan},
  {and} \bibinfo{person}{K.~S.~Rama Murty}.} \bibinfo{year}{2018}\natexlab{}.
\newblock \showarticletitle{Action {Recognition} {Based} on {Discriminative}
  {Embedding} of {Actions} {Using} {Siamese} {Networks}}. In
  \bibinfo{booktitle}{\emph{2018 25th {IEEE} {International} {Conference} on
  {Image} {Processing} ({ICIP})}}. \bibinfo{pages}{3473--3477}.
\newblock
\urldef\tempurl%
\url{https://doi.org/10.1109/ICIP.2018.8451226}
\showDOI{\tempurl}


\bibitem[\protect\citeauthoryear{Sakoe and Chiba}{Sakoe and Chiba}{1971}]%
        {SakoeChiba71}
\bibfield{author}{\bibinfo{person}{Hiroaki Sakoe} {and} \bibinfo{person}{Seibi
  Chiba}.} \bibinfo{year}{1971}\natexlab{}.
\newblock \showarticletitle{A Dynamic Programming Approach to Continuous Speech
  Recognition}. In \bibinfo{booktitle}{\emph{Proceedings of the Seventh
  International Congress on Acoustics, Budapest}}, Vol.~\bibinfo{volume}{3}.
  \bibinfo{publisher}{{Akad\'{e}miai} {Kiad\'{o}}},
  \bibinfo{address}{Budapest}, \bibinfo{pages}{65--69}.
\newblock


\bibitem[\protect\citeauthoryear{Sakoe and Chiba}{Sakoe and Chiba}{1978}]%
        {sakoe_dynamic_1978}
\bibfield{author}{\bibinfo{person}{H. Sakoe} {and} \bibinfo{person}{S. Chiba}.}
  \bibinfo{year}{1978}\natexlab{}.
\newblock \showarticletitle{Dynamic programming algorithm optimization for
  spoken word recognition}.
\newblock \bibinfo{journal}{\emph{IEEE Transactions on Acoustics, Speech, and
  Signal Processing}} \bibinfo{volume}{26}, \bibinfo{number}{1}
  (\bibinfo{date}{Feb.} \bibinfo{year}{1978}), \bibinfo{pages}{43--49}.
\newblock
\showISSN{0096-3518}
\urldef\tempurl%
\url{https://doi.org/10.1109/TASSP.1978.1163055}
\showDOI{\tempurl}


\bibitem[\protect\citeauthoryear{Salvador and Chan}{Salvador and Chan}{2007}]%
        {salvador_toward_2007}
\bibfield{author}{\bibinfo{person}{Stan Salvador} {and} \bibinfo{person}{Philip
  Chan}.} \bibinfo{year}{2007}\natexlab{}.
\newblock \showarticletitle{Toward {Accurate} {Dynamic} {Time} {Warping} in
  {Linear} {Time} and {Space}}.
\newblock \bibinfo{journal}{\emph{Intell. Data Anal.}} \bibinfo{volume}{11},
  \bibinfo{number}{5} (\bibinfo{date}{Oct.} \bibinfo{year}{2007}),
  \bibinfo{pages}{561--580}.
\newblock
\showISSN{1088-467X}
\urldef\tempurl%
\url{http://dl.acm.org/citation.cfm?id=1367985.1367993}
\showURL{%
\tempurl}


\bibitem[\protect\citeauthoryear{Serrà and Arcos}{Serrà and Arcos}{2014}]%
        {serra_empirical_2014}
\bibfield{author}{\bibinfo{person}{Joan Serrà} {and}
  \bibinfo{person}{Josep~Ll. Arcos}.} \bibinfo{year}{2014}\natexlab{}.
\newblock \showarticletitle{An empirical evaluation of similarity measures for
  time series classification}.
\newblock \bibinfo{journal}{\emph{Knowledge-Based Systems}}
  \bibinfo{volume}{67} (\bibinfo{date}{Sept.} \bibinfo{year}{2014}),
  \bibinfo{pages}{305--314}.
\newblock
\showISSN{0950-7051}
\urldef\tempurl%
\url{https://doi.org/10.1016/j.knosys.2014.04.035}
\showDOI{\tempurl}


\bibitem[\protect\citeauthoryear{Sun, Chen, Lee, and Liu}{Sun
  et~al\mbox{.}}{1992}]%
        {sun_time_1992}
\bibfield{author}{\bibinfo{person}{G.~Z. Sun}, \bibinfo{person}{H.~H. Chen},
  \bibinfo{person}{Y.~C. Lee}, {and} \bibinfo{person}{Y.~D. Liu}.}
  \bibinfo{year}{1992}\natexlab{}.
\newblock \showarticletitle{Time warping recurrent neural networks and
  trajectory classification}. In \bibinfo{booktitle}{\emph{[{Proceedings} 1992]
  {IJCNN} {International} {Joint} {Conference} on {Neural} {Networks}}},
  Vol.~\bibinfo{volume}{1}. \bibinfo{pages}{431--436 vol.1}.
\newblock
\urldef\tempurl%
\url{https://doi.org/10.1109/IJCNN.1992.287173}
\showDOI{\tempurl}


\bibitem[\protect\citeauthoryear{Tallec and Ollivier}{Tallec and
  Ollivier}{2018}]%
        {tallec_can_2018}
\bibfield{author}{\bibinfo{person}{Corentin Tallec} {and} \bibinfo{person}{Yann
  Ollivier}.} \bibinfo{year}{2018}\natexlab{}.
\newblock \showarticletitle{Can recurrent neural networks warp time?}
\newblock \bibinfo{journal}{\emph{International Conference on Learning
  Representations (ICLR)}} (\bibinfo{date}{Feb.} \bibinfo{year}{2018}).
\newblock


\bibitem[\protect\citeauthoryear{Wan, Wang, Hoi, Wu, Zhu, Zhang, and Li}{Wan
  et~al\mbox{.}}{2014}]%
        {wan_deep_2014}
\bibfield{author}{\bibinfo{person}{Ji Wan}, \bibinfo{person}{Dayong Wang},
  \bibinfo{person}{Steven Chu~Hong Hoi}, \bibinfo{person}{Pengcheng Wu},
  \bibinfo{person}{Jianke Zhu}, \bibinfo{person}{Yongdong Zhang}, {and}
  \bibinfo{person}{Jintao Li}.} \bibinfo{year}{2014}\natexlab{}.
\newblock \showarticletitle{Deep {Learning} for {Content}-{Based} {Image}
  {Retrieval}: {A} {Comprehensive} {Study}}. In
  \bibinfo{booktitle}{\emph{Proceedings of the 22Nd {ACM} {International}
  {Conference} on {Multimedia}}} \emph{(\bibinfo{series}{{MM} '14})}.
  \bibinfo{publisher}{ACM}, \bibinfo{address}{New York, NY, USA},
  \bibinfo{pages}{157--166}.
\newblock
\showISBNx{978-1-4503-3063-3}
\urldef\tempurl%
\url{https://doi.org/10.1145/2647868.2654948}
\showDOI{\tempurl}


\bibitem[\protect\citeauthoryear{Wang, Song, Leung, Rosenberg, Wang, Philbin,
  Chen, and Wu}{Wang et~al\mbox{.}}{2014}]%
        {wang_learning_2014}
\bibfield{author}{\bibinfo{person}{J. Wang}, \bibinfo{person}{Y. Song},
  \bibinfo{person}{T. Leung}, \bibinfo{person}{C. Rosenberg},
  \bibinfo{person}{J. Wang}, \bibinfo{person}{J. Philbin}, \bibinfo{person}{B.
  Chen}, {and} \bibinfo{person}{Y. Wu}.} \bibinfo{year}{2014}\natexlab{}.
\newblock \showarticletitle{Learning Fine-Grained Image Similarity with Deep
  Ranking}. In \bibinfo{booktitle}{\emph{2014 IEEE Conference on Computer
  Vision and Pattern Recognition}}. \bibinfo{pages}{1386--1393}.
\newblock
\showISSN{1063-6919}
\urldef\tempurl%
\url{https://doi.org/10.1109/CVPR.2014.180}
\showDOI{\tempurl}


\bibitem[\protect\citeauthoryear{Wang, Yan, and Oates}{Wang
  et~al\mbox{.}}{2017}]%
        {wang_time_2017}
\bibfield{author}{\bibinfo{person}{Z. Wang}, \bibinfo{person}{W. Yan}, {and}
  \bibinfo{person}{T. Oates}.} \bibinfo{year}{2017}\natexlab{}.
\newblock \showarticletitle{Time series classification from scratch with deep
  neural networks: {A} strong baseline}. In \bibinfo{booktitle}{\emph{2017
  {International} {Joint} {Conference} on {Neural} {Networks} ({IJCNN})}}.
  \bibinfo{pages}{1578--1585}.
\newblock


\bibitem[\protect\citeauthoryear{Warden}{Warden}{2017}]%
        {speechcommands}
\bibfield{author}{\bibinfo{person}{Pete Warden}.}
  \bibinfo{year}{2017}\natexlab{}.
\newblock \showarticletitle{Speech Commands: A public dataset for single-word
  speech recognition.}
\newblock \bibinfo{journal}{\emph{Dataset available from
  download.tensorflow.org/data/speech\_commands\_v0.01.tar.gz}}
  (\bibinfo{year}{2017}).
\newblock


\bibitem[\protect\citeauthoryear{Xing, Jordan, Russell, and Ng}{Xing
  et~al\mbox{.}}{2003}]%
        {xing2003distance}
\bibfield{author}{\bibinfo{person}{Eric~P Xing}, \bibinfo{person}{Michael~I
  Jordan}, \bibinfo{person}{Stuart~J Russell}, {and} \bibinfo{person}{Andrew~Y
  Ng}.} \bibinfo{year}{2003}\natexlab{}.
\newblock \showarticletitle{Distance metric learning with application to
  clustering with side-information}. In \bibinfo{booktitle}{\emph{Advances in
  neural information processing systems}}. \bibinfo{pages}{521--528}.
\newblock


\bibitem[\protect\citeauthoryear{Yeh, Zhu, Ulanova, Begum, Ding, Dau, Silva,
  Mueen, and Keogh}{Yeh et~al\mbox{.}}{2016}]%
        {7837992}
\bibfield{author}{\bibinfo{person}{C.~M. Yeh}, \bibinfo{person}{Y. Zhu},
  \bibinfo{person}{L. Ulanova}, \bibinfo{person}{N. Begum}, \bibinfo{person}{Y.
  Ding}, \bibinfo{person}{H.~A. Dau}, \bibinfo{person}{D.~F. Silva},
  \bibinfo{person}{A. Mueen}, {and} \bibinfo{person}{E. Keogh}.}
  \bibinfo{year}{2016}\natexlab{}.
\newblock \showarticletitle{Matrix Profile I: All Pairs Similarity Joins for
  Time Series: A Unifying View That Includes Motifs, Discords and Shapelets}.
  In \bibinfo{booktitle}{\emph{2016 IEEE 16th International Conference on Data
  Mining (ICDM)}}. \bibinfo{pages}{1317--1322}.
\newblock
\showISSN{2374-8486}
\urldef\tempurl%
\url{https://doi.org/10.1109/ICDM.2016.0179}
\showDOI{\tempurl}


\bibitem[\protect\citeauthoryear{Yi, Lei, Liao, and Li}{Yi
  et~al\mbox{.}}{2014}]%
        {yi_deep_2014}
\bibfield{author}{\bibinfo{person}{D. Yi}, \bibinfo{person}{Z. Lei},
  \bibinfo{person}{S. Liao}, {and} \bibinfo{person}{S.~Z. Li}.}
  \bibinfo{year}{2014}\natexlab{}.
\newblock \showarticletitle{Deep {Metric} {Learning} for {Person}
  {Re}-identification}. In \bibinfo{booktitle}{\emph{2014 22nd {International}
  {Conference} on {Pattern} {Recognition}}}. \bibinfo{pages}{34--39}.
\newblock
\urldef\tempurl%
\url{https://doi.org/10.1109/ICPR.2014.16}
\showDOI{\tempurl}


\bibitem[\protect\citeauthoryear{Zagoruyko and Komodakis}{Zagoruyko and
  Komodakis}{2015}]%
        {DBLP:conf/cvpr/ZagoruykoK15}
\bibfield{author}{\bibinfo{person}{Sergey Zagoruyko} {and}
  \bibinfo{person}{Nikos Komodakis}.} \bibinfo{year}{2015}\natexlab{}.
\newblock \showarticletitle{Learning to compare image patches via convolutional
  neural networks}. In \bibinfo{booktitle}{\emph{{IEEE} Conference on Computer
  Vision and Pattern Recognition, {CVPR} 2015, Boston, MA, USA, June 7-12,
  2015}}. \bibinfo{publisher}{{IEEE} Computer Society},
  \bibinfo{pages}{4353--4361}.
\newblock
\showISBNx{978-1-4673-6964-0}
\urldef\tempurl%
\url{https://doi.org/10.1109/CVPR.2015.7299064}
\showDOI{\tempurl}


\bibitem[\protect\citeauthoryear{Zhu, Zeng, Liao, Lei, Cai, and Zheng}{Zhu
  et~al\mbox{.}}{2018}]%
        {zhu_deep_2018}
\bibfield{author}{\bibinfo{person}{J. Zhu}, \bibinfo{person}{H. Zeng},
  \bibinfo{person}{S. Liao}, \bibinfo{person}{Z. Lei}, \bibinfo{person}{C.
  Cai}, {and} \bibinfo{person}{L. Zheng}.} \bibinfo{year}{2018}\natexlab{}.
\newblock \showarticletitle{Deep Hybrid Similarity Learning for Person
  Re-Identification}.
\newblock \bibinfo{journal}{\emph{IEEE Transactions on Circuits and Systems for
  Video Technology}} \bibinfo{volume}{28}, \bibinfo{number}{11}
  (\bibinfo{date}{Nov} \bibinfo{year}{2018}), \bibinfo{pages}{3183--3193}.
\newblock
\showISSN{1051-8215}


\end{thebibliography}

\end{document}